\pdfoutput=1
\documentclass[11pt, letterpaper, onecolumn, logo, copyright]{template_from_google}

\usepackage[T1]{fontenc}
\usepackage[utf8]{inputenc}

\usepackage{amsmath}
\usepackage{amsfonts}
\usepackage{amssymb}
\usepackage{amsthm}
\usepackage{bm}
\usepackage{nicefrac}

\usepackage{booktabs}
\usepackage{enumitem}
\usepackage{fancyhdr}
\usepackage{multirow}
\usepackage{tabularx}
\usepackage{multicol}
\usepackage{array}
\usepackage{longtable}
\usepackage{wrapfig}
\usepackage{authblk}

\usepackage{algorithm}
\usepackage[noend]{algpseudocode}

\usepackage{graphicx}
\usepackage{caption}
\usepackage{subcaption}
\usepackage{capt-of}

\usepackage[dvipsnames]{xcolor}
\usepackage{colortbl}

\usepackage{xspace}

\usepackage[numbers, sort&compress, square]{natbib}

\usepackage{hyperref}
\definecolor{darkblue}{rgb}{0.0, 0.0, 0.6}
\definecolor{darkred}{rgb}{0.7, 0.0, 0.0}
\hypersetup{
  pdffitwindow=true,
  pdfstartview={FitH},
  pdfnewwindow=true,
  colorlinks,
  linktocpage=true,
  linkcolor=darkred,
  urlcolor=darkblue,
  citecolor=darkblue
}
\usepackage[capitalise]{cleveref}
\crefname{equation}{Eq.}{Eqs.}

\newcommand{\Sec}[1]{Sec.~\ref{#1}}
\newcommand{\Fig}[1]{Fig.~\ref{#1}}
\newcommand{\Tab}[1]{Tab.~\ref{#1}}
\newcommand{\Append}[1]{Append.~\ref{#1}}

\title{Autoregressive Visual Generation Needs a Prologue}
\newcommand{\shorttitle}{Autoregressive Visual Generation Needs a Prologue}
\renewcommand{\headrulewidth}{0.4pt}

\reportnumber{}
\renewcommand{\today}{}

\author[1]{Bowen Zheng}
\author[2]{Weijian Luo}
\author[2]{Guang Yang}
\author[2]{Colin Zhang}
\author[1]{Tianyang Hu}

\affil[1]{The Chinese University of Hong Kong, Shenzhen}
\affil[2]{hi-Lab, Xiaohongshu Inc}

\begin{abstract}
    In this work, we propose \textbf{Prologue}, an approach to bridging the reconstruction--generation gap in autoregressive (AR) image generation. Instead of modifying visual tokens to satisfy both reconstruction and generation, Prologue generates a small set of \emph{prologue tokens} prepended to the visual token sequence. These prologue tokens are trained \emph{exclusively} with the AR cross-entropy (CE) loss, while visual tokens remain dedicated to reconstruction. This decoupled design lets us optimize generation through the AR model's true distribution without affecting reconstruction quality, which we further formalize from an ELBO perspective. On ImageNet 256$\times$256, Prologue-Base reduces gFID from 21.01 to 10.75 without classifier-free guidance while keeping reconstruction almost unchanged; Prologue-Large reaches a competitive rFID of 0.99 and gFID of 1.46 using a standard AR model without auxiliary semantic supervision. Interestingly, driven only by AR gradients, prologue tokens exhibit emergent semantic structure: linear probing on 16 prologue tokens reaches 35.88\% Top-1, far above the 23.71\% of the first 16 tokens from a standard tokenizer; resampling with fixed prologue tokens preserves a similar high-level semantic layout. Our results suggest a new direction: generation quality can be improved by introducing a separate learned generative representation while leaving the original representation intact.
\end{abstract}

\begin{document}

\maketitle

\setlength{\intextsep}{12pt plus 2pt minus 2pt}

\vspace{-10pt}

\begin{figure}[h]
    \centering
    \includegraphics[width=\linewidth]{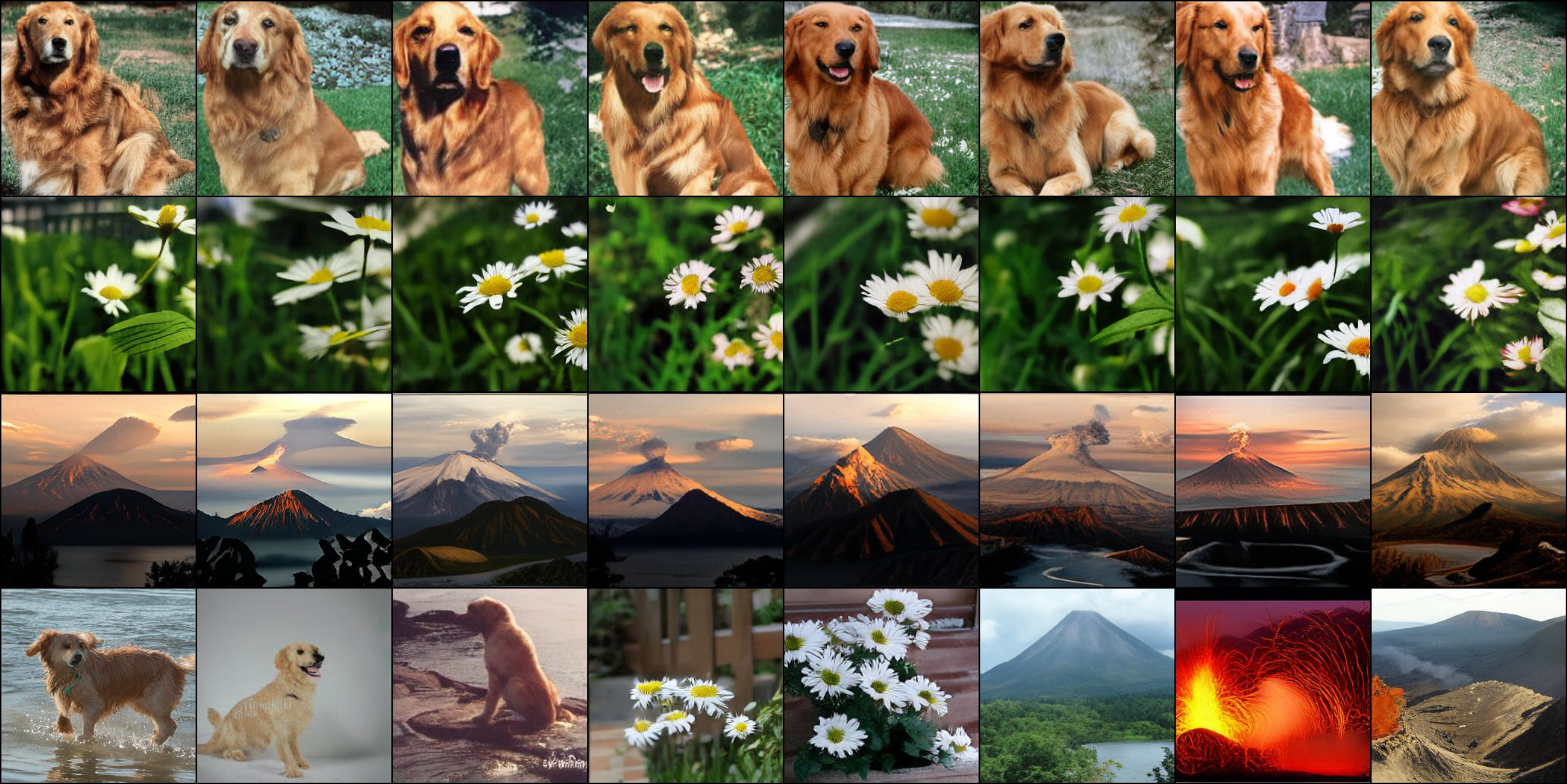}
    \vspace{-12pt}
    \caption{\textbf{Top three rows}: each row generated with \textbf{fixed prologue tokens} while resampling visual tokens. Samples within a row share remarkably consistent global structure, including viewpoint, spatial composition, color palette, and foreground/background style. This structure \textbf{emerges naturally} under the AR compression objective alone, \emph{without} any semantic supervision. \textbf{Bottom row}: samples drawn from \emph{different} prologue tokens, shown for reference.}
    \label{fig:disentangle}
\end{figure}

\section{Introduction}

Two-stage autoregressive (AR) image generation~\citep{Esser_2021_CVPR} can be viewed as a special case of the evidence lower bound (ELBO)~\citep{DBLP:journals/corr/KingmaW13}.
The ELBO contains a reconstruction term and a prior matching term.
Under deterministic encoding, the latter reduces to a negative log-prior term, whose dataset average is the cross-entropy between the aggregate posterior and the model prior.
This simplifies the training pipeline but introduces the well-known reconstruction-generation gap: the tokenizer never accounts for the learnability of the AR, and the data-model gap is reduced unilaterally by the AR.

Existing work explores several orthogonal directions to improve the AR-compatibility of tokens: semantic alignment~\citep{li2024imagefolderautoregressiveimagegeneration,xiong2025gigatokscalingvisualtokenizers,pmlr-v267-bachmann25a,zheng2025visionfoundationmodelseffective,Yao_2025_CVPR,yu2025representationalignmentgenerationtraining,zheng2025hitaholistictokenizerautoregressive}
, AR-aware encoding~\citep{pmlr-v267-bachmann25a,wang2025selftokdiscretevisualtokens,li2024imagefolderautoregressiveimagegeneration,miwa2025onedpieceimagetokenizermeets,zhang2026restoklearninghierarchicalresiduals,wu2026towards,zheng2025hitaholistictokenizerautoregressive}, hierarchical/coarse-to-fine generation~\citep{DBLP:conf/nips/TianJYPW24,ren2024flowarscalewiseautoregressiveimage,Wang_2025_CVPR,tang2024hartefficientvisualgeneration,huang2025spectralarspectralautoregressivevisual,liu2025detailflow1dcoarsetofineautoregressive,zhang2026restoklearninghierarchicalresiduals}.
But few of these works directly optimize the tokenizer to accommodate the AR model's actual distribution. The most direct way to achieve this is to jointly train an AR model and backpropagate its loss to the tokenizer.
However, AR and reconstruction gradients share the same set of variables $\mathbf{z}_v$, placing the two objectives on a single-variable Pareto front: increasing $\lambda$ degrades reconstruction, while decreasing $\lambda$ weakens the AR training signal (\Sec{sec:lambda_sweep}). This limits how much generation can improve.
We propose \textbf{Prologue}, which extends $\mathbf{z}_v$ to $[\mathbf{z}_p;\mathbf{z}_v]$, giving each objective its own variable: one group ($\mathbf{z}_p$) carries CE, while the other ($\mathbf{z}_v$) carries reconstruction.

Prologue is also grounded in a solid theoretical foundation. From the ELBO perspective, the prologue reduces prior matching from fitting the marginal $q(\mathbf{z}_v)$ to fitting the conditional $q(\mathbf{z}_v|\mathbf{z}_p)$, which is easier when $\mathbf{z}_p$ provides useful information. We further show that $\mathbf{z}_p$ admit a non-trivial optimum rather than collapsing to uninformative codes: since reconstruction objectives already preserve the information of visual tokens $\mathbf{z}_v$, collapsing $\mathbf{z}_p$ cannot improve reconstruction and provides little benefit to autoregressive prediction. To reduce AR loss, prologue tokens must instead capture information about $\mathbf{z}_v$ that is useful for autoregressive modeling. As a result, prologue tokens naturally evolve into compact generation-oriented representations that improve autoregressive prediction while leaving reconstruction quality unaffected (\Sec{sec:objective}).

Experimentally, Prologue significantly improves the quality of autoregressive visual generation, without relying on heuristic methods or auxiliary semantic regularization. On ImageNet 256$\times$256, Prologue-Base reduces the gFID w/o CFG from 21.01 to 10.75 ($\downarrow$48.8\%),
the gFID from 5.02 to 4.11 ($\downarrow$18.1\%), with reconstruction nearly unchanged (rFID 2.15 to 2.24) compared to the baseline. Scaling to Prologue-Large, it achieves competitive results with rFID of 0.99 and gFID of 1.46.
Prologue tokens, driven purely by AR gradients, develop semantic structure (linear probe 35.88\% vs.\ 23.71\%); resampling images with fixed prologue tokens show similar overall structure with different visual details (~\Fig{fig:disentangle}). A natural variant, Prologue-Post, learns the prologue from a frozen visual tokenizer. It keeps rFID unchanged, and still reduces gFID vs.\ the 2D baseline by 22.7\%: a prologue can be written together with the script, or composed separately after the script is complete.

\section{Prologue}
\label{sec:method}
\label{sec:framework}
\begin{figure}[t]
    \centering
    \includegraphics[width=\linewidth]{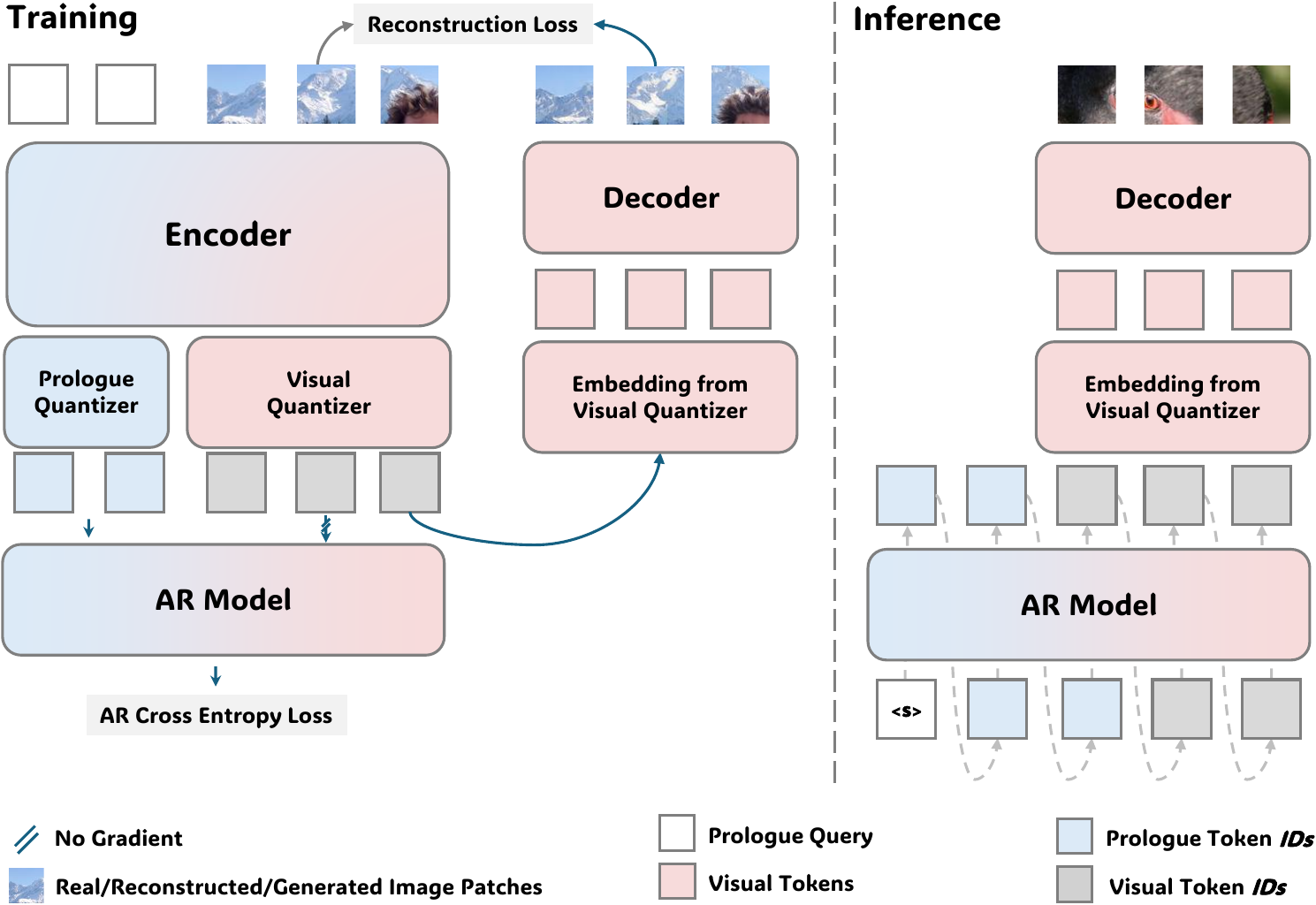}
    \caption{Prologue architecture. During training, a shared Encoder produces two groups of tokens: Visual Tokens (2D) handle reconstruction only, while Prologue Tokens (1D) receive AR gradients from the compact AR to reduce the data-model gap. The encoder is then frozen to train a full-size AR model (omitted for clarity). During inference, the full-size AR model conducts a standard autoregressive generation process. The model will first generate prologue tokens, then generate visual tokens based on prologue tokens. Finally, the visual tokens will be fed to the decoder to generate the final image. Blue lines represent the forward process with gradients; grey lines represent the forward pass without gradients.}
    \label{fig:architecture}
\end{figure}

As illustrated in \Fig{fig:architecture}, given an image $\mathbf{x}$, a shared Encoder $E$ simultaneously receives learnable queries and patch embeddings:
\begin{equation}
    [\mathbf{h}_p; \mathbf{h}_v] = E([\mathbf{q}_1, \ldots, \mathbf{q}_K; \text{Patchify}(\mathbf{x})])
\end{equation}
where $\mathbf{h}_p\in\mathbb{R}^{K\times d}$ corresponds to $K$ prologue positions (default $K\!=\!16$) and $\mathbf{h}_v\in\mathbb{R}^{N\times d}$ corresponds to $N\!=\!256$ visual positions. The Encoder is a bidirectional Transformer with global attention shared across all positions.

The two groups are quantized through separate quantizers and codebooks:
$\mathbf{h}_p$ is quantized by the Prologue Quantizer $\mathcal{Q}_p$ and Codebook $\mathcal{C}_p$ (size 1024) to yield $\mathbf{z}_p$,
and $\mathbf{h}_v$ by the Visual Quantizer $\mathcal{Q}_v$ and Codebook $\mathcal{C}_v$ (size 16384) to yield $\mathbf{z}_v$. Here $\mathbf{z}_p$ and $\mathbf{z}_v$ are discrete ids produced by quantizers.
Visual tokens are sent to the Decoder for image reconstruction: $\hat{\mathbf{x}}=D(\text{Lookup}(\mathcal{C}_v,\mathbf{z}_v))$, trained with $\ell_1$, perceptual~\citep{Zhang_2018_CVPR}, and adversarial losses~\citep{Isola_2017_CVPR}.
Prologue tokens do not participate in reconstruction.

The two groups are concatenated as $[\mathbf{z}_p;\mathbf{z}_v]$ and fed into a compact AR Model. The total loss is:
\begin{equation}
    \mathcal{L} = \underbrace{\mathcal{L}_\text{recon}(\mathbf{x}, \hat{\mathbf{x}})}_{\text{acts on visual tokens only}} + \lambda\cdot\underbrace{\mathcal{L}_\text{AR}(\mathbf{z}_p, \mathbf{z}_v)}_{\text{acts on prologue tokens only}}
\end{equation}
where $\mathcal{L}_\text{AR}$ is the autoregressive CE loss over the full sequence $[\mathbf{z}_p;\mathbf{z}_v]$.
Prob-STE (\Sec{sec:details}) routes the gradient of $\mathcal{L}_\text{AR}$ to prologue tokens only; visual tokens do not directly receive AR gradients.
Although AR gradients can indirectly influence visual representations through bidirectional attention updates, this indirect pathway is far weaker than the direct gradient route (rFID rises by only 0.09 in \Sec{sec:lambda_sweep}). Prologue-Post (\Sec{sec:prologue_variants}) freezes the entire visual pathway for complete isolation.
From the perspective of visual tokens, the AR no longer needs to fit the difficult marginal distribution $q(\mathbf{z}_v)$ directly,
but instead fits the simpler conditional distribution $q(\mathbf{z}_v|\mathbf{z}_p)$ under prologue conditioning.

\subsection{Theoretical analysis from the Evidence Lower Bound (ELBO) perspective}
\label{sec:objective}

As noted in the introduction, two-stage training is a special case of the ELBO under deterministic encoding (VQ). Let $q_\phi$ denote the encoder, $p_\psi$ the decoder, and $p_\theta$ the AR prior. The VQ deterministic encoding $q_\phi(\mathbf{z}|\mathbf{x})=\delta(\mathbf{z}-\mathrm{Enc}_\phi(\mathbf{x}))$ reduces the ELBO to:
\begin{equation}
    \text{ELBO} = \underbrace{\log p_\psi(\mathbf{x}|\mathbf{z}^*)}_{\text{Stage\,1: reconstruction}} + \underbrace{\log p_\theta(\mathbf{z}^*)}_{\text{Stage\,2: AR}}
\label{eq:elbo_deterministic}
\end{equation}
Two-stage training decouples these terms: $q_\phi$ is shaped for reconstruction alone and never considers the AR.
Taking expectation over the dataset, the Stage\,2 objective amounts to minimizing the cross-entropy $H(q_\phi, p_\theta) = H(q_\phi) + D_\text{KL}(q_\phi\|p_\theta)$.
Optimizing $\theta$ can only shrink the $D_\text{KL}$ term, while the structure of $q_\phi$ is determined entirely by Stage\,1, which is the origin of the reconstruction-generation gap. We refer to $D_\text{KL}(q_\phi\|p_\theta)$ as the \textbf{data-model gap} hereafter.

\paragraph{ELBO of Prologue.}
Prologue augments the latent to $\mathbf{z}=[\mathbf{z}_p;\mathbf{z}_v]$, where $\mathbf{z}_p$ is an auxiliary latent that does not participate in reconstruction ($p_\psi(\mathbf{x}|\mathbf{z}_p,\mathbf{z}_v)=p_\psi(\mathbf{x}|\mathbf{z}_v)$) and only assists AR modeling through a structured prior. Under deterministic encoding:
\begin{equation}
    \text{ELBO}_\text{Prologue}
    = \underbrace{\log p_\psi(\mathbf{x}|\mathbf{z}_v^*)}_{\text{reconstruction (same)}}
    + \underbrace{\log p_\theta(\mathbf{z}_p^*)}_{\text{prologue prior}}
    + \underbrace{\log p_\theta(\mathbf{z}_v^*|\mathbf{z}_p^*)}_{\text{conditional visual prior}}
\label{eq:elbo_prologue}
\end{equation}
Compared with Eq.~\ref{eq:elbo_deterministic}, the reconstruction term is identical, while the prior term changes from $\log p_\theta(\mathbf{z}_v^*)$ to $\log p_\theta(\mathbf{z}_p^*)+\log p_\theta(\mathbf{z}_v^*|\mathbf{z}_p^*)$.
In other words, the AR's prior matching shifts from fitting the marginal $q(\mathbf{z}_v)$ to fitting the conditional $q(\mathbf{z}_v|\mathbf{z}_p)$. Without altering $q(\mathbf{z}_v)$, the inequality $H(\mathbf{z}_v|\mathbf{z}_p)\leq H(\mathbf{z}_v)$ ensures that conditional matching is strictly easier, provided that $\mathbf{z}_p$ carries non-trivial information about $\mathbf{z}_v$, i.e., $I(\mathbf{z}_p;\mathbf{z}_v)>0$. We show below that AR gradients naturally prevent $\mathbf{z}_p$ from collapsing, as degenerate prologue tokens yield zero benefit over the two-stage baseline.
\begin{figure}[t]
    \centering
    \includegraphics[width=\linewidth]{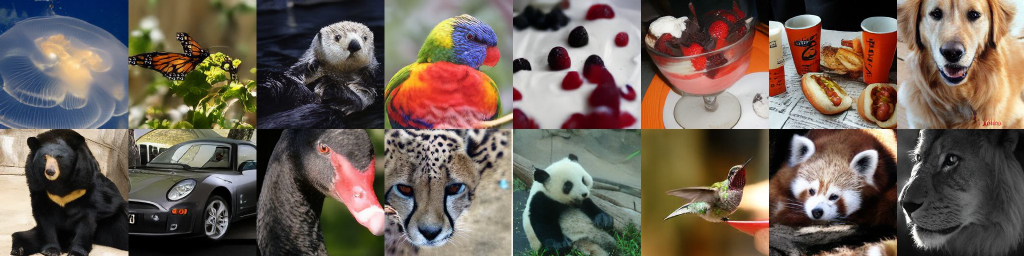}
    \vspace{-4pt}
    \caption{Samples from Prologue L--XL (semantic cfg=0.7, visual cosine cfg: scale=2.25, p=0.225).}
    \label{fig:samples}
\end{figure}

\paragraph{Prologue avoid collapse pathway.}
The loss $\mathcal{L}_\text{recon}+\lambda\mathcal{L}_\text{AR}$ Prologue optimizes is
the weighted negative ELBO of Eq.~\ref{eq:elbo_prologue}; gradient routing sends
AR gradients only to $\mathbf{z}_p$, leaving $q(\mathbf{z}_v)$ intact and
reconstruction nearly unaffected (\Sec{sec:lambda_sweep}). Within the AR loss term,
$H(\mathbf{z}_v)$ is held fixed by reconstruction, so the only
$\phi_p$-dependent contributions decompose as
$H(\mathbf{z}_p)-I(\mathbf{z}_p;\mathbf{z}_v)$ plus a residual conditional KL.
Collapsing $\mathbf{z}_p$ to a constant zeros $H(\mathbf{z}_p)$ and
$I(\mathbf{z}_p;\mathbf{z}_v)$ simultaneously and yields no decrease in the
loss and is not a descent direction. The only way to lower the AR loss is
to grow $I(\mathbf{z}_p;\mathbf{z}_v)$ faster than $H(\mathbf{z}_p)$, i.e., to
compress information from $q(\mathbf{z}_v)$ into $q(\mathbf{z}_p)$ in a form
the AR can predict autoregressively. Prologue tokens therefore must encode
useful, AR-friendly information by construction
(full derivation in \Append{app:theory}).

\subsection{Implementation details}
\label{sec:details}

\paragraph{Prob-STE.}
\label{sec:probste}
The AR CE loss is defined on discrete token indices. Standard STE performs straight-through estimation in the embedding space and is ineffective for discrete CE.
Prob-STE first computes the soft assignment $p_i=\text{softmax}(\mathbf{h}_p\cdot\mathbf{c}_i/\tau)$;
the forward pass uses $\arg\max$ to produce a one-hot vector, while the backward pass propagates gradients through $\tilde{p}_i=p_i+\text{sg}(\hat{p}_i-p_i)$.
This is equivalent to the straight-through variant of Gumbel-Softmax~\citep{DBLP:conf/iclr/JangGP17,DBLP:conf/iclr/MaddisonMT17} without Gumbel noise, an idea originating from an open-source implementation~\citep{lucidrains2022vq} and further scaled by IBQ~\citep{shi2025scalableimagetokenizationindex}.
On the AR Model side, the input embedding for prologue tokens is computed as $\tilde{\mathbf{p}}\cdot\mathbf{W}_\text{emb}$ rather than a discrete lookup,
so AR gradients can flow back to the Encoder along the soft probability (see more details in \Append{app:prob-ste}).

\paragraph{Visual token dropout.}
During Stage 1 training~\ref{sec:setup}, all visual tokens in the AR input are replaced with \texttt{[EOS]} with probability $p_\text{drop}\!=\!0.5$.
The visual part in Prologue contains 256 tokens, compared with only 16 prologue tokens, and the local dependencies among visual tokens are strong.
The AR model may rely heavily on these visual tokens and ignore the prologue prefix, preventing prologue tokens from receiving useful gradients.
When all visual tokens are dropped, the AR model is forced to rely on prologue tokens for prediction, providing stronger gradient signals to the prologue.
$p_\text{drop}\!=\!0.5$ strikes the best balance (\Tab{tab:ablation}).

\paragraph{CFG for Prologue.}
At inference time, different classifier-free guidance~\citep{DBLP:journals/corr/abs-2207-12598} scales are applied to the two groups.
Prologue tokens encode class-level information, so an excessively large scale reduces intra-class diversity;
visual tokens encode spatial details and require stronger guidance.
In practice, $s_\text{pro}\approx 0.75$ to $0.9$ and $s_\text{vis}\approx 2.0$ to $2.5$.

\subsection{Prologue-Post: plug-and-play upgrade for existing Tokenizers.}
\label{sec:prologue_variants}
As prologue tokens carry only AR gradients while visual tokens carry only reconstruction gradients, the two objectives are structurally decoupled by design, making the framework agnostic to \emph{when} and \emph{how} the prologue is introduced. At one end of this spectrum, we instantiate \textbf{Prologue-Post}, which attaches the prologue to an existing Tokenizer in a post-hoc manner: all parameters of the existing Tokenizer (Tokenizer, Decoder) are frozen, while only an additional prologue tokenizer $(\mathcal{E}_p,\mathcal{Q}_p)$,  and a compact AR Model are newly introduced and trained with $\mathcal{L}_\text{AR}$. As a result, given any off-the-shelf Visual Tokenizer, one can prepend a small set of prologue tokens and train only the lightweight prologue components to improve downstream AR generation (gFID $\downarrow$22.7\% vs.\ the 2D baseline, \Tab{tab:ablation}), without retraining or modifying the original Tokenizer. We validate this variant on the Base configuration as a proof of concept (\Sec{sec:ablation}, \Append{app:ablation}), and leave large-scale exploration to future work.

\section{Experiments}
\label{sec:experiments}

\subsection{Setup}
\label{sec:setup}
We conduct our main experiments on ImageNet~\citep{DBLP:conf/cvpr/DengDSLL009} 256$\times$256. We evaluate reconstruction results with rFID~\citep{DBLP:conf/nips/HeuselRUNH17} and PSNR, and class-conditional generation results with gFID, Inception Score (IS)~\citep{DBLP:conf/nips/SalimansGZCRCC16}, Precision, and Recall. rFID is computed on the ImageNet validation set and gFID follows the ADM~\citep{DBLP:conf/nips/DhariwalN21} protocol. The Tokenizer has B and L variants; the AR Model has B, L and X-Large variants. Prologue L--XL means L Size tokenizer with XL Size ar model. We adopt a two stage training pipeline following previous works. In Stage~1, the Tokenizer trains for 150/200 (B/L) epochs with a compact AR (40M) supplying AR gradients to prologue tokens (\Sec{sec:framework}).
In Stage~2, a full-size AR trains from scratch on the frozen tokenizer for 400/800 (AR paired with B/L Tokenizer) epochs. We also test a full one-stage training variant \textbf{Prologue-OneStage} but observe degraded results (\Append{app:ablation}). For more experimental details, see \Append{app:details}.

\subsection{Controlled comparison}
\label{sec:main_results}

\setlength{\intextsep}{0pt}
\vspace{-0.72\baselineskip}
\begin{wraptable}[10]{r}{0.48\textwidth}
\centering
\small
\setlength{\abovecaptionskip}{0pt}
\setlength{\belowcaptionskip}{0pt}
\caption{Controlled comparison on ImageNet 256$\times$256 (B--B; 86M Tokenizer, 86M Decoder, and 115M AR). Prologue uses $\lambda\!=\!3$. We report the best gFID for each model.}\label{tab:main}
\footnotesize
\renewcommand{\tabcolsep}{2pt}
\renewcommand{\arraystretch}{0.84}
\resizebox{\linewidth}{!}{%
\begin{tabular}{lcccc}
\toprule
Method & rFID $\downarrow$ & PSNR $\uparrow$ & gFID $\downarrow$ & gFID w/o CFG $\downarrow$ \\
\midrule
1D Tokenizer & 2.11 & 20.71 & 6.10 & 19.32 \\
2D Tokenizer & 2.15 & 20.71 & 5.02 & 21.01 \\
\rowcolor{gray!15}
Prologue & 2.24 & 20.64 & 4.11 & 10.75 \\
Prologue-Post & 2.15 & 20.71 & 3.88 & 11.04 \\
\bottomrule
\end{tabular}%
}
\vspace{-7pt}
\end{wraptable}\vspace{-0.2\baselineskip}
\Tab{tab:main} compares the standard two-stage baseline with Prologue under
the same experiment settings. Prologue significantly improves generation quality:
gFID w/o CFG drops from 21.01 to 10.75 ($\downarrow$48.8\%) and
gFID from 5.02 to 4.11 ($\downarrow$18.1\%), while reconstruction is almost the same (rFID 2.15/2.24, PSNR 20.71/20.64). Notably, Prologue-Post
improves generation without affecting reconstruction at all. These results are
consistent with our theoretical analysis (\Sec{sec:objective}): prologue tokens
can improve generation without compromising reconstruction.

\setlength{\intextsep}{12pt plus 2pt minus 2pt}
\subsection{Comprehensive Comparison}

\Tab{tab:sota} presents detailed reconstruction and generation comparisons. Prologue achieves competitive results against state-of-the-art methods. Specifically, Prologue-Base matches classical 2D-token AR baselines such as LlamaGen in compression rate and model size while improving generation quality (gFID 4.11 vs.\ 5.46 for LlamaGen-B). Prologue-Large adopts a compression rate and tokenizer size comparable to modern 1D-token AR works such as AliTok, and reaches comparable performance without their auxiliary training techniques (e.g., dual perceptual losses or periodic codebook clustering). Our best model, Prologue L--XL, reaches rFID 0.99 and gFID 1.46, on par with leading diffusion models, masked-prediction methods, and non-standard AR variants. We emphasize that Prologue is trained \emph{without} the auxiliary techniques commonly used in recent tokenizers, including semantic alignment to DINO/SigLIP features, periodic codebook clustering, tail/nested token dropout, diffusion-based decoders, or distillation/initialization from pre-trained model. The tokenizer is trained from scratch with reconstruction, adversarial, and perceptual losses, and the generator is trained with a simple next-token prediction task.

\subsection{Scalability}
\label{sec:scalability}
The bottom of \Tab{tab:sota} shows that Prologue scales well along both
axes: the tokenizer and the AR model. Scaling the tokenizer from Base to Large reduces rFID from 2.24 to 0.99 and, under the same AR size, reduces
gFID w/o CFG from 10.75 (B--B) to 5.02 (L--B). Scaling the AR model with the Base tokenizer, gFID w/o CFG
drops from 10.75 (B--B) to 6.56 (B--L) and 5.22 (B--XL). With the Large tokenizer, gFID w/o CFG drops
from 5.02 (L--B) to 2.81 (L--L) and 2.26 (L--XL). These results show consistent gains from scaling both the tokenizer and the AR model.

\begin{figure}[t]
    \centering
    \includegraphics[width=\linewidth]{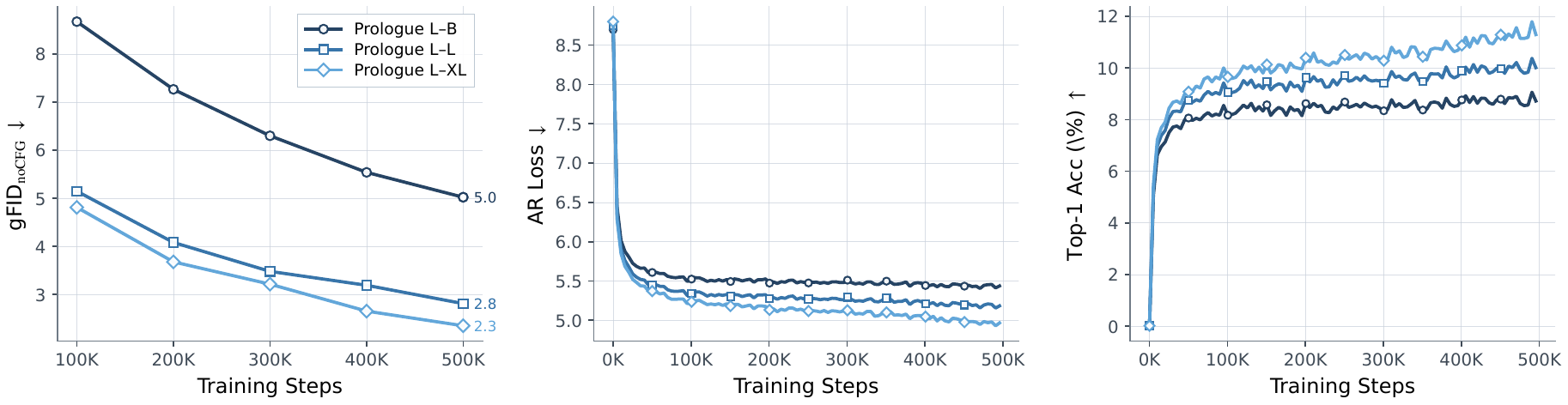}
    \caption{Training dynamics for Prologue L-series. \textbf{Left:} gFID w/o CFG. \textbf{Middle:} AR training loss. \textbf{Right:} Top-1 accuracy. As the AR model scales, loss decreases, accuracy rises, and gFID w/o CFG improves throughout training.}
    \label{fig:training_curve}
\end{figure}

\begin{table}[t]
\centering
\caption{Comparison of reconstruction and class-conditional generation on ImageNet 256$\times$256.  $\dag$: Uses auxiliary semantic alignment to DINO/SigLIP features. \textbf{$\ddag$: This method uses an additional periodic codebook/token clustering, which requires trichy parameter tuning.} $\S$: Uses tail/nested/token dropout. $\diamond$: Uses a diffusion/flow-matching decoder on discrete tokens. $\star$: Distilled/Initialized from a pre-trained tokenizer. Prologue achieves competitive results without relying on any of these techniques.}
\label{tab:sota}
\resizebox{\linewidth}{!}{%
\begin{tabular}{lllllllllll}
\toprule
\multicolumn{1}{l|}{\multirow{2}{*}{Method}}                                          & \multicolumn{4}{l|}{Tokenizer}                                                & \multicolumn{6}{l}{Generator}                                                                  \\ \cmidrule(l){2-11}
\multicolumn{1}{l|}{}                                                                 & \#Param(Enc+Dec).    & Codebook & rFID↓  & \multicolumn{1}{l|}{PSNR↑}  & \#Param. & gFID↓ & gFID\textsubscript{w/o CFG} ↓ & IS↑   & Pre.↑ & Rec.↑ \\ \midrule
\multicolumn{11}{c}{\textit{Continuous Tokens}}                                                                                                                                                                                          \\ \midrule
\multicolumn{1}{l|}{LDM-4-G~\citep{Rombach_2022_CVPR}}                                       & 55M  & -        & 0.27   & \multicolumn{1}{l|}{27.53}  & 400M     & 3.60  & -             & 247.7 & 0.87  & 0.48  \\
\multicolumn{1}{l|}{DiT-XL/2~\citep{Peebles_2023_ICCV}}                                      & 84M  & -        & 0.62   & \multicolumn{1}{l|}{-}      & 675M     & 2.27  & 9.62          & 278.2 & 0.83  & 0.57  \\
\multicolumn{1}{l|}{MAR-H~\citep{DBLP:conf/nips/LiTLDH24}}                                         & 70M  & -        & 0.53   & \multicolumn{1}{l|}{-}      & 943M     & 1.55  & 2.35          & 303.7 & 0.81  & 0.62  \\
\multicolumn{1}{l|}{REPA-SiT-XL$\dag$~\citep{yu2025representationalignmentgenerationtraining}}                              & 84M  & -        & 0.62   & \multicolumn{1}{l|}{-}      & 675M     & 1.42  & 5.90          & 305.7 & 0.80  & 0.65  \\
\multicolumn{1}{l|}{LightningDiT-XL$\dag$~\citep{Yao_2025_CVPR}}                  & 70M  & -        & 0.28   & \multicolumn{1}{l|}{-}      & 675M     & 1.35  & 2.17          & 295.3 & 0.79  & 0.65  \\ \midrule
\multicolumn{11}{c}{\textit{Discrete Tokens + Non-standard AR / Non-AR}}                                                                                                                                                                 \\ \midrule
\multicolumn{1}{l|}{MAGVIT-v2~\citep{yu2024language}}                                        & -          & 262K     & -      & \multicolumn{1}{l|}{-}      & 307M     & 1.78  & 3.65          & 319.4 & -     & -     \\
\multicolumn{1}{l|}{VAR-$d$30~\citep{DBLP:conf/nips/TianJYPW24}}                                            & -          & 4K       & -      & \multicolumn{1}{l|}{-}      & 2.0B     & 1.92  & -             & 323.1 & 0.82  & 0.59  \\
\multicolumn{1}{l|}{TiTok-B-64$\star$~\citep{yu2024imageworth32tokens}}                                        & 86+86M     & 4K       & 1.70   & \multicolumn{1}{l|}{-}      & 177M     & 2.48  & 3.08         & 214.7 & -     & -     \\
\multicolumn{1}{l|}{TiTok-S-128$\star$~\citep{yu2024imageworth32tokens}}                                        & 22+22M     & 4K       & 1.71   & \multicolumn{1}{l|}{-}      & 287M     & 1.97  & 4.44          & 281.8 & -     & -     \\
\multicolumn{1}{l|}{ImageFolder$^{\star\dag\S}$~\citep{li2024imagefolderautoregressiveimagegeneration}}                       & 86+86M   & 4K$\times$2 & 0.80 & \multicolumn{1}{l|}{-}     & 362M     & 2.60  & -             & 295.0 & 0.75  & 0.63  \\
\multicolumn{1}{l|}{RAR-XXL~\citep{yu2024randomizedautoregressivevisualgeneration}}                                              & 36+36M     & 1K      & 2.28   & \multicolumn{1}{l|}{-}      & 1.5B     & 1.48  & -             & 326.0 & 0.80  & 0.63  \\
\multicolumn{1}{l|}{ResTok$^{\dag\S}$~\citep{zhang2026restoklearninghierarchicalresiduals}}                                 & 662M       & 8K       & 1.28   & \multicolumn{1}{l|}{-}      & 326M     & 2.34  & 4.56             & 257.8 & 0.79  & 0.60  \\ \midrule
\multicolumn{11}{c}{\textit{Discrete Tokens + Standard AR}}                                                                                                                                                                              \\ \midrule
\multicolumn{1}{l|}{LlamaGen-B~\citep{sun2024autoregressivemodelbeatsdiffusion}}                                      & 36+36M     & 16K      & 2.19   & \multicolumn{1}{l|}{-}      & 111M     & 5.46  & 26.26             & 193.6 & 0.83  & 0.45  \\
\multicolumn{1}{l|}{LlamaGen-XXL~\citep{sun2024autoregressivemodelbeatsdiffusion}}                                    & 36+36M     & 16K      & 2.19   & \multicolumn{1}{l|}{-}      & 1.4B     & 2.34  & 14.64             & 253.9 & 0.80  & 0.59  \\
\multicolumn{1}{l|}{FlexTok d18-d28$^{\dag\S\diamond}$~\citep{pmlr-v267-bachmann25a}}                             & 1.4B & 64K     & 1.45   & \multicolumn{1}{l|}{13.96}      & 1.4B    & 1.71  &          -    & 284.9    & 0.82     & 0.61     \\
\multicolumn{1}{l|}{GigaTok-B-L$^{\dag}$~\citep{xiong2025gigatokscalingvisualtokenizers}}                           & 622M       & 16K      & 0.81   & \multicolumn{1}{l|}{21.21}      & 1.4B     & 2.03  & -             &  238.5   & 0.80     & 0.63    \\
\multicolumn{1}{l|}{VFMTok-SigLIP2-XXL$^{\dag}$~\citep{zheng2025visionfoundationmodelseffective}}                                   & - & 16K     & 0.94   & \multicolumn{1}{l|}{-}      & 1.4B     & 2.16  & 1.98          & 272.0 & 0.83  & 0.60  \\
\multicolumn{1}{l|}{AliTok-B$^{\ddag}$~\citep{wu2026towards}}                                & 390M       & 4K       & 0.86   & \multicolumn{1}{l|}{20.97}  & 177M     & 1.44  & 2.40          & 319.5 & 0.77  & 0.65  \\
\multicolumn{1}{l|}{AliTok-XL$^{\ddag}$~\citep{wu2026towards}}                               & 390M       & 4K       & 0.86   & \multicolumn{1}{l|}{20.97}  & 662M     & 1.28  & 1.88          & 306.3 & 0.79  & 0.65  \\\midrule
\multicolumn{11}{c}{\textbf{Prologue (Ours, Discrete Tokens + Standard AR)}}                                                                                                                                                                   \\ \midrule
\multicolumn{1}{l|}{Prologue B--B}                                & 86+86M     & 1K+16K   & 2.24   & \multicolumn{1}{l|}{20.64}  & 115M     & 4.11  & 10.75         & 210.3 & 0.83  & 0.48  \\
\multicolumn{1}{l|}{Prologue B--L}                                & 86+86M     & 1K+16K   & 2.24   & \multicolumn{1}{l|}{20.64}  & 305M     & 2.67  & 6.56          & 251.2 & 0.82  & 0.56  \\
\multicolumn{1}{l|}{Prologue B--XL}                               & 86+86M     & 1K+16K   & 2.24   & \multicolumn{1}{l|}{20.64}  & 685M     & 2.43  & 5.22          & 252.6 & 0.80  & 0.59  \\
\multicolumn{1}{l|}{Prologue L--B}                                & 86+305M & 1K+4K    & 0.99   & \multicolumn{1}{l|}{20.01}  & 115M     & 2.15  & 5.02          & 219.9 & 0.79  & 0.60  \\
\multicolumn{1}{l|}{Prologue L--L}                                & 86+305M & 1K+4K    & 0.99   & \multicolumn{1}{l|}{20.01}  & 305M     & 1.52  & 2.81          & 251.6 & 0.77  & 0.66  \\
\multicolumn{1}{l|}{Prologue L--XL}                               & 86+305M & 1K+4K    & 0.99   & \multicolumn{1}{l|}{20.01}  & 685M     & 1.46  & 2.26          & 257.7 & 0.78  & 0.66  \\ \bottomrule
\end{tabular}%
}
\end{table}
\vspace{3pt}

\section{Analysis}
\label{sec:analysis}

\subsection{Why Prologue works: \texorpdfstring{$\lambda$}{lambda} sweep and gradient analysis}
\label{sec:lambda_sweep}
\label{sec:why_2d_analysis}
\label{sec:why_prologue}
\setlength{\intextsep}{0pt}
\vspace{-0.68\baselineskip}
\noindent
\begin{wrapfigure}[13]{r}{0.48\textwidth}
\centering
\small
\includegraphics[width=\linewidth]{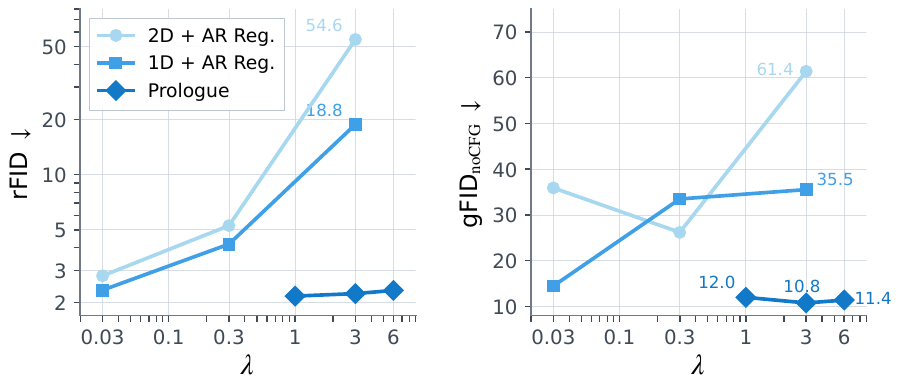}
\vspace{-14pt}
\setlength{\abovecaptionskip}{0pt}
\setlength{\belowcaptionskip}{0pt}
\caption{$\lambda$ sweep comparison between 2D + AR Reg., 1D + AR Reg., and Prologue.
\textbf{Left:} rFID vs.\ $\lambda$; \textbf{right:} gFID w/o CFG
vs.\ $\lambda$. All models are trained with the B--B configuration with 86M
Tokenizer, 86M Decoder, and 115M AR.}\label{fig:lambda_sweep}
\vspace{-8pt}
\end{wrapfigure}%
When AR gradients flow directly to visual tokens, $\nabla_\phi\mathcal{L}_\text{AR}$ includes $\nabla H(\mathbf{z}_v)$: the gradient can reduce CE by lowering the data entropy rather than improving AR fit. This phenomenon is observed across modalities~\citep{wang2025larptokenizingvideoslearned}.
Even with a very small $\lambda$, CE still alters the intrinsic structure of $\mathbf{z}_v$ (\Sec{sec:attention}).
Prologue structurally blocks this pathway: AR gradients cannot directly reach
$H(\mathbf{z}_v)$. Collapsing $\mathbf{z}_p$ to a constant drives both
$H(\mathbf{z}_p)$ and $I(\mathbf{z}_p;\mathbf{z}_v)$ to zero, thus providing no
benefit for the objective. The only effective optimization direction is for
$\mathbf{z}_p$ to encode information that enables autoregressive prediction of
$\mathbf{z}_v$ under the AR model, naturally forming an information bottleneck
(full derivation in \Append{app:theory}).

\Fig{fig:lambda_sweep} verifies this analysis by sweeping $\lambda$ under
three AR routing strategies. When reconstruction and AR gradients share
$\mathbf{z}_v$, the two objectives trade off against each other and the
trade-off worsens with $\lambda$. For reconstruction, rFID of both 1D and 2D tokenizers collapses at moderate
$\lambda$ (18.8 for 1D, 54.6 for 2D at $\lambda\!=\!3$). Even $\lambda\!=\!0.3$
already alters reconstruction, and only at $\lambda\!=\!0.03$ does rFID stay
close to the two-stage baseline and Prologue (but still worse). For generation, the 1D tokenizer
benefits from joint AR training at small $\lambda$, but its best
gFID w/o CFG (14.45) still lags Prologue (10.75). The 2D
tokenizer behaves differently: joint AR training produces worse generation than
the two-stage baseline. Our attention analysis (\Sec{sec:attention}) shows
that forcing 2D tokens into the 1D AR raster order destroys their original
local attention pattern and makes AR fitting harder.

These results suggest that sharing a single latent for both reconstruction and
generation couples the two objectives on a single-variable Pareto front,
limiting how much generation can improve. After latent augmentation, the trade-off disappears: rFID fluctuates by only 0.16 as $\lambda$
varies from 1 to 6; gFID w/o CFG reaches its minimum of 10.75 at
$\lambda\!=\!3$ (11.98 at $\lambda\!=\!1$, 11.38 at $\lambda\!=\!6$). Notably,
Prologue operates at $\lambda$ values (1--6) far above those feasible for
direct AR ($\leq\!0.03$), as the structural decoupling removes the
reconstruction constraint on $\lambda$.

\setlength{\intextsep}{12pt plus 2pt minus 2pt}
\subsection{Attention patterns: the structural advantage of 2D tokens}
\label{sec:attention}
\begin{figure}[t]
    \centering
    \includegraphics[width=\linewidth]{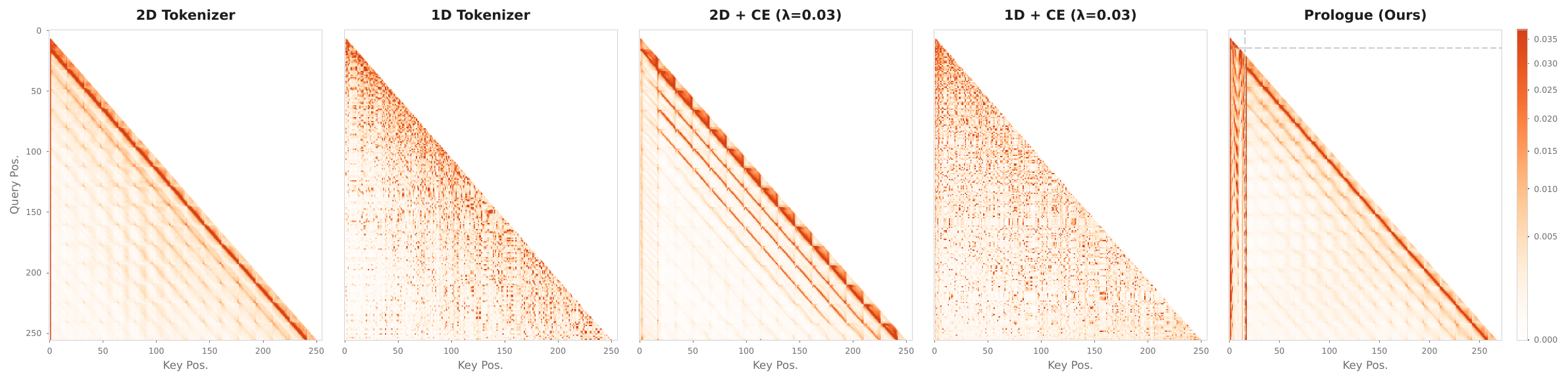}
    \caption{AR attention patterns (Layer\,7; self-attention diagonal and first token masked). (a)~2D Tokenizer shows a clear local pattern (left neighbor + above). (b)~1D Tokenizer lacks spatial locality; attention is dispersed. (c)~CE on 2D visual tokens ($\lambda\!=\!0.03$) severely disrupts the local structure as the 2D spatial order inherently conflicts with the 1D AR raster order, and AR gradients destroy the spatial locality. (d)~CE on 1D visual tokens ($\lambda\!=\!0.03$) makes the attention pattern more uniform compared to the 1D baseline. (e)~Prologue preserves the 2D local pattern while introducing prologue attention (vertical stripes; gray dashed line marks the prologue/visual boundary), showing that the AR model learns to use prologue tokens' global information to assist local visual token prediction. Full heatmaps across all layers are provided in Appendix \Fig{fig:attn_all_layers}.}
    \label{fig:attention}
\end{figure}

\setlength{\intextsep}{0pt}
\Tab{tab:attention_compare} and \Fig{fig:attention} compare 2D and 1D tokens.
The 2D baseline's attention naturally exhibits a local pattern, with each token primarily attending to its left and top neighbors (\Fig{fig:attention}a); this is a surprising intrinsic advantage of the 2D raster order.
1D tokens, produced by a bidirectional Encoder, lack this structure (\Fig{fig:attention}b), making AR fitting harder (gFID of 1D: 6.10 vs.\ 2D: 5.02, \Tab{tab:attention_compare}).

When AR gradients are applied to 2D visual tokens (\Fig{fig:attention}c), the local structure is severely disrupted.
The root cause is a \emph{fundamental conflict} between the 2D spatial order encoded by the Tokenizer and the 1D raster order required by the AR model: AR gradients push visual tokens toward easier next-token prediction along the raster scan, destroying the spatial locality that made 2D tokens advantageous in the first place.
Interestingly, this conflict is less severe for 1D tokens (\Fig{fig:attention}d): Since 1D tokens carry no strong inherent ordering, AR gradients (or other
supervision signals) can shape their order without overriding a pre-existing
structure. This is an advantage of 1D tokenizers.

\begin{wraptable}[12]{r}{0.45\textwidth}
\vspace{-0.6\baselineskip}
\centering
\small
\setlength{\abovecaptionskip}{0pt}
\setlength{\belowcaptionskip}{8pt}
\caption{Comparison between 2D and 1D tokens. Under the same Encoder and AR sizes, 2D tokens possess natural spatial locality, making AR fitting easier and yielding better reconstruction and generation results. All models are trained under the B--B configuration with 86M Tokenizer, 86M Decoder, 115M AR, and without AR regularization.}\label{tab:attention_compare}\label{tab:2d_vs_1d}
\renewcommand{\tabcolsep}{2pt}
\resizebox{\linewidth}{!}{%
\begin{tabular}{lccccc}
\toprule
Token structure & rFID $\downarrow$ & PSNR $\uparrow$ & AR Loss $\downarrow$ & gFID $\downarrow$ & gFID w/o CFG $\downarrow$ \\
\midrule
1D Tokenizer & 2.11 & 20.71 & 7.89 & 6.10 & 19.32 \\
2D Tokenizer & 2.15 & 20.71 & 7.73 & 5.02 & 21.01 \\
\bottomrule
\end{tabular}%
}
\end{wraptable}In Prologue, we use 1D tokens for the prologue tokens and 2D tokens for visual content to leverage the advantages of each in their corresponding role. The visual tokens in Prologue do not directly receive AR gradients, so the 2D local structure is fully preserved (\Fig{fig:attention}). A new attention pattern also emerges: visual tokens attend heavily to the prologue prefix (vertical stripes on the left),
showing that the AR Model indeed learns to use the global information from prologue tokens to assist local visual token prediction (\Fig{fig:attention}e).

\setlength{\intextsep}{12pt plus 2pt minus 2pt}

\subsection{Ablation study}
\label{sec:ablation}

We ablate Prologue design space under the B--B configuration: visual dropout $p_\text{drop}$, prologue token length $K$, prologue codebook size $|\mathcal{C}_p|$, Prob-STE temperature $\tau$, AR Loss weight $\lambda$, Prologue-Post, and Prologue-OneStage. Visual dropout is the most sensitive; $K$ strongly affects gFID, while codebook size and $\tau$ are relatively stable. Full results and discussion are in \Append{app:ablation} (\Tab{tab:ablation}).

\subsection{Emergent semantic representations and visualization}

\Tab{tab:linear_probe} uses linear probing on ImageNet classification
(following MAE's protocol) to investigate what information AR gradients drive
prologue tokens to encode. We mean-pool the first 16 tokens produced by the
Prologue tokenizer and the standard 2D tokenizer, and train a linear classifier
on top. Prologue tokens reach 35.88\% Top-1 accuracy, well above the 23.71\% of
the first 16 tokens from the 2D baseline ($\uparrow$12.17 pp). This semantic
content is driven entirely by the AR gradients, confirming the information
bottleneck analysis (\Sec{sec:why_prologue}): prologue tokens naturally
develop a semantic summary of the image under a pure generation objective.

\setlength{\intextsep}{0pt}
\vspace{-0.7\baselineskip}
\begin{wraptable}[10]{r}{0.48\textwidth}
\centering
\small
\setlength{\abovecaptionskip}{0pt}
\setlength{\belowcaptionskip}{2pt}
\caption{Linear probing results on ImageNet Classification task using only the \emph{first 16 tokens} of each tokenizer. Driven solely by AR gradients, Prologue's 16 prologue tokens show better performance compared to the first 16 tokens of a standard 2D tokenizer.}\label{tab:linear_probe}
\footnotesize
\renewcommand{\tabcolsep}{2.5pt}
\renewcommand{\arraystretch}{0.92}
\begin{tabular}{lcc}
\toprule
Method & Top-1 $\uparrow$ & Top-5 $\uparrow$ \\
\midrule
2D Tokenizer & 23.71 & 44.17 \\
Prologue & \textbf{35.88} & \textbf{60.28} \\
\bottomrule
\end{tabular}
\vspace{-5pt}
\end{wraptable}

\begin{figure}[t]
    \centering
    \includegraphics[width=\linewidth]{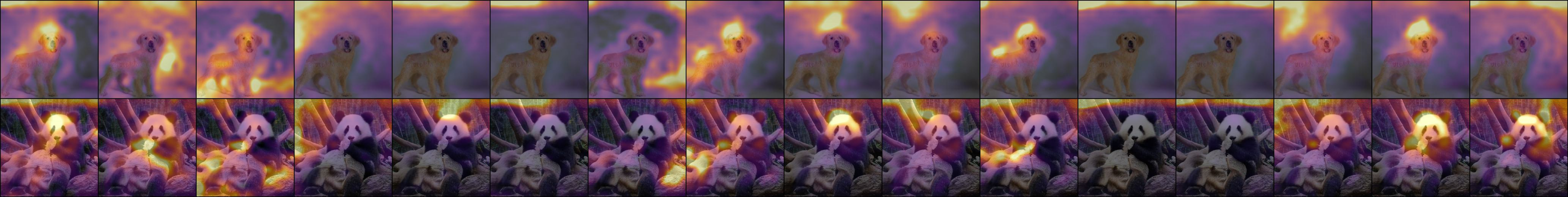}
    \vspace{-4pt}
    \caption{Attention heatmaps from each prologue token to visual tokens (averaged from the last 8 layers and heads). The prologue tokens show rich semantic structure that provides information to visual tokens for objects and background at different regions, indicating a learned spatial division of labor.}
    \label{fig:attn_heatmap}
\end{figure}

\Fig{fig:disentangle} illustrates the information division between the two groups through controlled variation.
With prologue tokens held fixed and only visual tokens changed during generation, the class and overall layout remain constant while only texture and local details vary.
\Fig{fig:attn_heatmap} further visualizes the per-token attention: different prologue tokens specialize in different spatial regions (e.g., foreground object vs.\ background), confirming that each token captures a distinct aspect of the global structure.
This division of labor is not designed through explicit losses but arises naturally from the information bottleneck under an AR compression objective.

\section{Discussion}
\label{sec:discussion}
\label{sec:comparison}
Prologue aims to supervise the tokenizer with the AR model's distribution. Existing approaches explore several directions to improve the AR-compatibility of tokens. One line aligns tokenizer features with pretrained vision models that carry dense semantic information, such as the DINO series~\citep{oquab2024dinov, simeoni2025dinov3}, to inject semantic structure into the tokens~\citep{li2024imagefolderautoregressiveimagegeneration,xiong2025gigatokscalingvisualtokenizers,pmlr-v267-bachmann25a,zheng2025visionfoundationmodelseffective,Yao_2025_CVPR,yu2025representationalignmentgenerationtraining,zheng2025hitaholistictokenizerautoregressive}. Another line introduces AR-aware regularization on the tokenizer, including causal decoders~\citep{wu2026towards}, tail-dropout objectives that implicitly induce sequential structure~\citep{pmlr-v267-bachmann25a,wang2025selftokdiscretevisualtokens,li2024imagefolderautoregressiveimagegeneration,miwa2025onedpieceimagetokenizermeets,zhang2026restoklearninghierarchicalresiduals}, and flow-matching decoders~\citep{wang2025selftokdiscretevisualtokens,pmlr-v267-bachmann25a,Sargent_2025_ICCV}. A third line designs hierarchical or coarse-to-fine tokenizers whose tokens align better with the AR prior~\citep{DBLP:conf/nips/TianJYPW24,ren2024flowarscalewiseautoregressiveimage,Wang_2025_CVPR,tang2024hartefficientvisualgeneration,huang2025spectralarspectralautoregressivevisual,liu2025detailflow1dcoarsetofineautoregressive,zhang2026restoklearninghierarchicalresiduals}. The most direct route is to backpropagate an AR loss to the tokenizer~\citep{wang2025larptokenizingvideoslearned, ramanujan2025worsebetternavigatingcompressiongeneration}, but existing methods rely on very small AR weights or weaker surrogates (e.g., MSE) due to the entropy collapse problem we analyze in \Sec{sec:why_prologue}, which limits performance.

Prologue takes a different path: AR gradients are routed only to a
dedicated set of prologue tokens, leaving the visual tokens unaffected. This
avoids the collapse of visual tokens observed in prior direct-AR-loss
approaches~\citep{wang2025larptokenizingvideoslearned, ramanujan2025worsebetternavigatingcompressiongeneration}, and lets us close the data-model gap with the
actual AR likelihood while leaving reconstruction untouched. One could instead try to optimize $D_\text{KL}(q_\phi\|p_\theta)$ directly.
However, $H(q_\phi)\!=\!\mathbb{E}_{q_\phi}[\log q_\phi]$ is a functional of
the entire data distribution and cannot be computed per-sample for a
deterministic encoder. Kernel discrepancy is feasible but depends on kernel choice and does not directly exploit the autoregressive structure of $p_\theta$; score-based objectives require additional derivation in the discrete space. The cross-entropy naturally
matches the per-token prediction structure of $p_\theta$ without additional design, making it the simplest viable path.

In the AR setting, Prologue realizes a form of learned conditioning through
sequence prefixing: it does not alter the distribution of visual tokens but
learns an additional prefix that reduces AR modeling difficulty. This idea can
in principle extend to other generation paradigms (masked prediction,
diffusion), though different conditioning mechanisms and gradient pathways
would be needed; we leave this to future work.

Despite Prologue's theoretical grounding and strong empirical performance,
this work has several limitations. We validate only on ImageNet 256$\times$256;
larger-scale datasets and higher resolutions remain to be explored. The
prologue token count $K\!=\!16$ and codebook size 1024 work well in the current
setting, but different datasets may require adjustment; the ablation
(\Sec{sec:ablation}) shows that $K$ has a large effect on gFID. The 16
prologue tokens add roughly 6\% to the AR sequence length. Finally, only CE
has been verified as the optimization objective; the effect under other
prior-matching objectives (e.g., kernel-based ~\citep{JMLR:v13:gretton12a} or score-based~\citep{DBLP:conf/nips/LuoHZSLZ23} distributional matching) remains
open.

\section{Conclusion}
In this work, we propose \textbf{Prologue}, an approach to bridge the reconstruction-generation gap in autoregressive (AR) visual generation. During training, Prologue learns to generate a small set of prologue tokens prepended to the visual tokens; the prologue tokens are trained exclusively with the AR loss, while the visual tokens remain dedicated to the reconstruction objective. During generation, Prologue first generates the prologue tokens, then generates the visual tokens conditioned on them. Formally, this design conditions the AR prior as $p_\theta(\mathbf{z}_p)\,p_\theta(\mathbf{z}_v\mid\mathbf{z}_p)$, reducing modeling difficulty while protecting reconstruction quality. Prologue-Base reduces gFID w/o CFG from 21.01 to 10.75 ($\downarrow$48.8\%) and gFID from 5.02 to 4.11 ($\downarrow$18.1\%), with reconstruction nearly unchanged. Scaling to Prologue-Large, it achieves competitive results with rFID 0.99 and gFID 1.46 without additional tricks. Prologue tokens also naturally develop semantic structure under pure AR gradients: 16 tokens reach 35.88\% Top-1 on linear probing, well above the 23.71\% of a standard tokenizer. More broadly, this design principle of providing a dedicated latent variable for optimizing the data-model prior while keeping the original representation untouched can be applied to any system where a reconstruction-generation gap exists.

{\small
\bibliography{references}
}

\appendix
\renewcommand{\thetable}{A\arabic{table}}
\renewcommand{\theHtable}{A\arabic{table}}
\renewcommand{\thefigure}{A\arabic{figure}}
\renewcommand{\theHfigure}{A\arabic{figure}}
\setcounter{table}{0}
\setcounter{figure}{0}

\section{Broader impacts}
\label{app:impacts}
Our work may positively impact research on efficient and controllable visual generation. The core idea of improving generative quality by decomposing prior matching into an easier conditional modeling problem can be applied to methods beyond discrete and autoregressive generation. This framework provides a way to improve generation quality without perturbing or altering the original data, thereby avoiding potential information loss and raising the upper bound of generative quality.
Improvements in image generation quality may also introduce misuse risks, including misleading synthetic imagery, impersonation, and disinformation. We therefore recommend that any release of checkpoints or generated samples be accompanied by clear usage guidelines, model cards, provenance documentation, and, where appropriate, watermarking or detection tools.

\section{Related work}
\label{app:related}

\paragraph{Discrete image tokenization.}
The paradigm of discrete image tokens was established by VQ-VAE~\citep{DBLP:conf/nips/OordVK17}; VQGAN~\citep{Esser_2021_CVPR} introduced adversarial training and markedly improved reconstruction quality.
Subsequent work has improved codebook utilization and efficiency through lookup-free quantization (LFQ)~\citep{yu2024language}, finite scalar quantization (FSQ)~\citep{mentzer2024finite}, residual quantization (RQ)~\citep{Lee_2022_CVPR}.
On the reconstruction frontier, improved VQGAN variants~\citep{DBLP:conf/iclr/YuLKZPQKXBW22,zhu2023designingbetterasymmetricvqgan} and mode-seeking diffusion autoencoders (FlowMo)~\citep{Sargent_2025_ICCV} have pushed rFID substantially lower.
On the token structure side, TiTok~\citep{yu2024imageworth32tokens} introduced a compact 1D tokenizer using learnable queries, FlexTok~\citep{pmlr-v267-bachmann25a} supports variable-length encoding, and One-D-Piece~\citep{miwa2025onedpieceimagetokenizermeets} explores quality-controllable compression.
ImageFolder~\citep{li2024imagefolderautoregressiveimagegeneration} uses semantic-detail dual-branch product quantization, and HART~\citep{tang2024hartefficientvisualgeneration} decomposes a continuous latent into discrete and continuous parts.

\paragraph{Bridging the tokenizer--AR gap.}
A central challenge in two-stage pipelines is that tokens optimized for reconstruction may be unnecessarily difficult for the AR prior to model.
Existing work explores several orthogonal directions to improve AR-compatibility.
\textit{Semantic alignment:}
GigaTok~\citep{xiong2025gigatokscalingvisualtokenizers} aligns decoder features with frozen DINOv2~\citep{oquab2024dinov};
ResTok~\citep{zhang2026restoklearninghierarchicalresiduals} matches DINOv3~\citep{simeoni2025dinov3} features at both encoder and decoder;
VFMTok~\citep{zheng2025visionfoundationmodelseffective} builds the entire encoder on top of frozen vision foundation models;
ImageFolder~\citep{li2024imagefolderautoregressiveimagegeneration} uses DINOv2 contrastive losses in a dual-branch tokenizer.
\textit{AR-aware encoding:}
AliTok~\citep{wu2026towards} trains a causal decoder to align with the AR generation order;
tail-dropout objectives, first explored in neural audio codecs~\citep{DBLP:journals/taslp/ZeghidourLOST22} and later adapted to visual tokenizers~\citep{miwa2025onedpieceimagetokenizermeets,pmlr-v267-bachmann25a,li2024imagefolderautoregressiveimagegeneration,wang2025selftokdiscretevisualtokens}, implicitly induce sequential structure by requiring reconstruction from token prefixes.
\textit{Hierarchical and coarse-to-fine generation:}
VAR~\citep{DBLP:conf/nips/TianJYPW24} predicts at progressively finer scales,
FlowAR~\citep{ren2024flowarscalewiseautoregressiveimage} integrates flow matching across scales,
ResTok~\citep{zhang2026restoklearninghierarchicalresiduals} uses hierarchical residuals so that lower-level tokens carry global information,
and PAR~\citep{Wang_2025_CVPR} parallelizes generation across spatially distant tokens.
\textit{Direct generative-loss backpropagation:}
The most natural approach is to backpropagate the AR loss directly to the tokenizer~\citep{wang2025larptokenizingvideoslearned,ramanujan2025worsebetternavigatingcompressiongeneration}.
However, this objective captures only the likelihood term of the ELBO; neglecting the entropy term $H(q_\phi(\mathbf{z}|\mathbf{x}))$ removes the penalty for data entropy collapse, easily driving the tokenizer toward degenerate solutions.
Existing mitigations rely on reduced loss weights, surrogate MSE objectives, limiting how much generation can improve (\Sec{sec:why_prologue}).
Prologue retains CE as the optimization objective but routes the gradients to dedicated prologue tokens, structurally decoupling the two objectives. It is orthogonal to the semantic alignment and order-aware directions above and can in principle be combined with them.

\paragraph{Autoregressive visual generation.}
Autoregressive image generation dates back to PixelCNN-style decoders~\citep{oord2016conditionalimagegenerationpixelcnn} and has progressed to Transformer-based systems~\citep{pmlr-v80-parmar18a,pmlr-v119-chen20s,pmlr-v139-ramesh21a,yu2022scaling}.
Building on VQ-based AR pipelines~\citep{DBLP:conf/nips/OordVK17,Esser_2021_CVPR}, LlamaGen~\citep{sun2024autoregressivemodelbeatsdiffusion} demonstrated that standard LLM-style backbones can scale autoregressive image generation to competitive quality.
Beyond raster-order next-token prediction, recent work explores diverse generation paradigms:
VAR~\citep{DBLP:conf/nips/TianJYPW24} uses next-scale prediction,
FlowAR~\citep{ren2024flowarscalewiseautoregressiveimage} integrates flow matching,
HART~\citep{tang2024hartefficientvisualgeneration} couples discrete AR with a residual diffusion module,
DetailFlow~\citep{liu2025detailflow1dcoarsetofineautoregressive} and SpectralAR~\citep{huang2025spectralarspectralautoregressivevisual} re-order tokens along resolution or spectral axes,
RandAR~\citep{Pang_2025_CVPR} and RAR~\citep{yu2024randomizedautoregressivevisualgeneration} randomize the generation order.
On the continuous side, MAR~\citep{DBLP:conf/nips/LiTLDH24} removes vector quantization entirely and uses a diffusion head for per-token generation.

\paragraph{Relation to prefix tuning and prompt tuning.}
Prologue prepends a learned token prefix to the visual sequence, which bears a
superficial resemblance to prefix tuning~\citep{li-liang-2021-prefix} and prompt
tuning~\citep{lester-etal-2021-power} in NLP. However, the two ideas differ
fundamentally: prefix/prompt tuning attaches \emph{static} learned vectors that
are shared across all inputs and independent of any specific instance, serving
as a task-level conditioning mechanism. In contrast, Prologue learns a
\emph{tokenizer} that extracts instance-specific information from each image
and compresses it into a compact prefix optimized for autoregressive
prediction. In other words, prefix tuning conditions on a task; Prologue
conditions on the data.

\section{Why prologue tokens do not collapse: full derivation}
\label{app:theory}
Prologue transforms the AR modeling objective from marginal matching $p_\theta(\mathbf{z}_v)\approx q(\mathbf{z}_v)$ to conditional matching $p_\theta(\mathbf{z}_v|\mathbf{z}_p)\approx q(\mathbf{z}_v|\mathbf{z}_p)$. Below we derive the mathematical properties of this transformation starting from the CE of the joint distribution. Let $q_\phi(\mathbf{z}_p,\mathbf{z}_v)$ denote the joint data prior induced by the Encoder, and $p_\theta(\mathbf{z}_p,\mathbf{z}_v)$ the joint model prior from the AR. The joint CE decomposes as:
\begin{equation}
    \mathcal{L}_\text{CE} = H(q_\phi(\mathbf{z}_p,\mathbf{z}_v)) + D_\text{KL}(q_\phi(\mathbf{z}_p,\mathbf{z}_v) \| p_\theta(\mathbf{z}_p,\mathbf{z}_v))
\end{equation}
The data prior entropy decomposes as:
\begin{equation}
    H(\mathbf{z}_p,\mathbf{z}_v) = H(\mathbf{z}_v) + H(\mathbf{z}_p|\mathbf{z}_v) = H(\mathbf{z}_v) + H(\mathbf{z}_p) - I(\mathbf{z}_p;\mathbf{z}_v)
\end{equation}
$H(\mathbf{z}_v)$ is maintained independently by $\mathcal{L}_\text{recon}$ (visual tokens do not directly receive AR gradients; the indirect effect from shared attention parameters is negligible in experiments, with rFID rising by only 0.09) and is not affected by $\phi_p$. The CE gradient's optimization of $\phi_p$ (prologue encoder) is approximately equivalent to:
\begin{equation}
    \min_{\phi_p}\big[H(\mathbf{z}_p) - I(\mathbf{z}_p;\mathbf{z}_v) + D_\text{KL}(q_\phi(\mathbf{z}_p)\|p_\theta(\mathbf{z}_p)) + \mathbb{E}_{q(\mathbf{z}_p)}[D_\text{KL}(q_\phi(\mathbf{z}_v|\mathbf{z}_p)\|p_\theta(\mathbf{z}_v|\mathbf{z}_p))]\big]
\end{equation}
\paragraph{Collapse pathway analysis.} If $\mathbf{z}_p$ degenerates to a constant ($H(\mathbf{z}_p)\to 0$): $I(\mathbf{z}_p;\mathbf{z}_v)\to 0$, $D_\text{KL}(q(\mathbf{z}_p)\|p_\theta(\mathbf{z}_p))\to 0$, and the conditional $D_\text{KL}$ reverts to the unconditional $D_\text{KL}(q(\mathbf{z}_v)\|p_\theta(\mathbf{z}_v))$. The total CE then equals $H(\mathbf{z}_v)+D_\text{KL}(q(\mathbf{z}_v)\|p_\theta(\mathbf{z}_v))$, identical to the two-stage baseline. Collapsing prologue tokens provides zero benefit.
\paragraph{Effective pathway analysis.} Non-degenerate $\mathbf{z}_p$ reduces $H(\mathbf{z}_v|\mathbf{z}_p)$ and the conditional $D_\text{KL}$ by increasing $I(\mathbf{z}_p;\mathbf{z}_v)$. The increase in $H(\mathbf{z}_p)$ incurs a small cost, but because prologue tokens are compact (16 tokens, 1K codebook), this cost is far smaller than the benefit from conditioning. AR gradients therefore naturally drive prologue tokens to encode global information that helps predict visual tokens, forming an information bottleneck.
\paragraph{Contrast: why direct optimization of visual tokens collapses.} If AR gradients flow directly to $\phi_v$ (visual encoder), $\nabla_{\phi_v}\mathcal{L}_\text{CE}$ contains $\nabla_{\phi_v}H(\mathbf{z}_v)$, and the gradient can push $H(\mathbf{z}_v)$ to 0 (codebook degenerates to a constant). This is an effective CE shortcut: the loss drops substantially, but reconstruction collapses. Prologue structurally blocks this shortcut: AR gradients cannot directly reach $H(\mathbf{z}_v)$.

\section{Ablation study}
\label{app:ablation}

\begin{table}[h]
\centering
\caption{Ablation study (B--B). Default configuration \colorbox{gray!15}{highlighted}.}
\label{tab:ablation}
\small
\begin{tabular}{llccc}
\toprule
Factor & Value & rFID $\downarrow$ & gFID w/o CFG $\downarrow$ & gFID $\downarrow$ \\
\midrule
\rowcolor{gray!15}
Default & -- & 2.24 & 10.75 & 4.11 \\
\midrule
\multirow{2}{*}{Prologue codebook} & 256 & 2.22 & 12.10 & 4.36 \\
 & 4096 & 2.25 & 12.68 & 4.36 \\
\midrule
\multirow{2}{*}{Prologue token count} & 4 & 2.24 & 13.49 & 4.20 \\
 & 64 & 2.25 & 12.31 & 6.81 \\
\midrule
\multirow{2}{*}{Prob-STE $\tau$} & 0.01 & 2.18 & 11.66 & 4.02 \\
 & 1.0 & 2.14 & 11.49 & 4.08 \\
\midrule
\multirow{2}{*}{AR weight $\lambda$} & 1.0 & 2.17 & 11.98 & 4.21 \\
 & 6.0 & 2.33 & 11.38 & 4.28 \\
\midrule
\multirow{2}{*}{Visual dropout $p$} & 0.0 & 2.16 & 12.32 & 4.28 \\
 & 1.0 & 2.25 & 12.67 & 4.53 \\
\midrule
\multicolumn{5}{l}{\textit{Training Prologue with Pre-trained Visual Tokenizer}} \\
Prologue-Post & -- & 2.15 & 11.04 & 3.88 \\
\midrule
\multicolumn{5}{l}{\textit{Training Prologue with OneStage}} \\
Prologue-OneStage & -- & 2.09 & 21.00 & 5.41  \\
\bottomrule
\end{tabular}
\end{table}

\Tab{tab:ablation} varies one factor at a time from the default ($p_\text{drop}\!=\!0.5$, $K\!=\!16$, $|\mathcal{C}_p|\!=\!1024$, $\tau\!=\!0.1$, $\lambda\!=\!3.0$).

\paragraph{Visual dropout $p$.} The most sensitive hyperparameter. At $p\!=\!0.0$, the AR Model can drive CE very low using only the local patterns of visual tokens, drowning out the prologue gradient signal; gFID reaches only 4.28. At $p\!=\!1.0$, prologue tokens receive the strongest gradient, but the AR Model never sees visual tokens during training. $p\!=\!0.5$ is the best balance.

\paragraph{Prologue token count $K$ and codebook size.} At $K\!=\!4$, the bottleneck is too tight and gFID reaches 4.20; at $K\!=\!64$, the sequence is too long and gFID degrades to 6.81. Codebook size is relatively insensitive: from 256 to 4096, gFID is 4.36 in both cases.

\paragraph{Prob-STE $\tau$ and $\lambda$.} $\tau$ is insensitive within a reasonable range (0.01 to 1.0, gFID 4.02 to 4.08). $\lambda$ from 1.0 to 6.0 causes rFID to rise by only 0.16, whereas CE directly on visual tokens at comparable $\lambda$ severely degrades reconstruction (\Fig{fig:lambda_sweep}).

\paragraph{Prologue-Post.} As described in \Sec{sec:prologue_variants}, Prologue-Post freezes the entire pre-trained visual pathway and introduces only the prologue components. rFID stays exactly at 2.15, with gFID 3.88 and gFID w/o CFG 11.04 under the same search protocols as \Tab{tab:main}; joint training in the default row trades a higher gFID (4.11) for rFID 2.24 with shared Encoder updates.

\paragraph{Prologue-OneStage.}
We also evaluate a simplified one-stage training pipeline in which the full-size
AR model is trained jointly with the tokenizer, eliminating the separate
Stage~2. We observe that this leads to degraded generation quality compared to
two-stage training, which is expected since the tokenizer's latent distribution
shifts continuously during joint optimization and is not determined until the
tokenizer is frozen.

\paragraph{Large Tokenizer variants and end-to-end generation quality.}
Recent tokenizers have explored stronger LPIPS backbones: MaskBit~\citep{weber2024maskbit} uses ResNet-50, while TA-TiTok~\citep{kim2025democratizingtexttoimagemaskedgenerative} and AliTok~\citep{wu2026towards} use ConvNeXt-S (alongside VGG in AliTok). Following this trend, we replace VGG16 with a standalone ConvNeXt-S in our larger tokenizer. As shown in Table~\ref{tab:large_variants}, this reduces rFID from 1.88 to 0.99 ($\downarrow$47.3\%) at 4K codebook size, while VGG16 retains an edge on PSNR, indicating that the two losses optimize along different axes. Expanding the codebook from 4K to 16K further lowers rFID to 0.92, but gFID slightly worsens (2.15 to 2.30), suggesting larger codebooks improve reconstruction precision at the cost of AR modeling difficulty.
Prologue-Base retains VGG to match classical baselines such as LlamaGen, while Prologue-Large adopts ConvNeXt-S to match modern tokenizers such as AliTok. The improvement from ConvNeXt-S is orthogonal to Prologue itself: Table~\ref{tab:prologue_vs_baseline} shows Prologue provides consistent gains at both backbones.
\begin{table}[h]
\centering
\caption{Effect of perceptual loss backbone and codebook size (L--B). All configurations use the same architecture and training hyperparameters ($\lambda\!=\!3$,
$\tau\!=\!0.1$,
$p_\text{drop}\!=\!0.5$). A better perceptual loss backbone improves the reconstruction and generation quality. Reducing the codebook from 16K to 4K slightly drops the reconstruction results but improves the generation.}
\label{tab:large_variants}
\label{tab:tok_e2e}
\small
\begin{tabular}{llccccccc}
\toprule
Perceptual loss & CB & BPP & rFID $\downarrow$ & PSNR $\uparrow$ & gFID w/o CFG $\downarrow$ & gFID $\downarrow$ \\
\midrule
VGG16      & 4K  & 0.049 & 1.88 & \textbf{20.45} &  10.87 & 4.56 \\
ConvNeXt-S & 4K  & 0.049 & 0.99 & 20.01 & 5.02 &  \textbf{2.15} \\
ConvNeXt-S & 16K & 0.057 & \textbf{0.92} & 20.36 &  6.73  & 2.30 \\
\bottomrule
\end{tabular}
\end{table}

\paragraph{Prologue improvement over the 2D baseline.}
\Tab{tab:prologue_vs_baseline} compares the standard 2D Tokenizer and Prologue at two scales, where the only variable is whether prologue tokens are introduced. All other factors (Tokenizer size, perceptual loss, codebook, AR Model, training steps) are identical within each group.

\begin{table}[h]
\centering
\caption{Prologue improvement over the 2D baseline. The only variable within each group is whether prologue tokens are introduced; all other factors are identical.}
\label{tab:prologue_vs_baseline}
\small
\begin{tabular}{llccccc}
\toprule
Scale & Method & rFID $\downarrow$ & gFID w/o CFG $\downarrow$ & $\Delta$ & gFID $\downarrow$ & $\Delta$ \\
\midrule
\multirow{2}{*}{B--B} & 2D Baseline & 2.15 & 21.01 & -- & 5.02 & -- \\
 & Prologue & 2.24 & 10.75 & $\downarrow$48.8\% & 4.11 & $\downarrow$18.1\% \\
\midrule
\multirow{2}{*}{L--B} & 2D Baseline & 0.99 & 7.54 & -- & 2.59 & -- \\
 & Prologue & 0.99 & 5.02 & $\downarrow$33.5\% & 2.15 & $\downarrow$17.0\% \\
\bottomrule
\end{tabular}
\end{table}

\section{Implementation details}
\label{app:details}

\paragraph{Computational resources.}
All experiments are conducted on a cluster of NVIDIA GPUs. We report the estimated training cost converted to an A100 setting. For Stage~1, the Base and Large tokenizers require approximately 30 and 44 hours on 8 A100 GPUs, respectively. For Stage~2, the Base and Large AR models require approximately 13 and 25 hours on 8 A100 GPUs, respectively, while the XL AR model requires approximately 25 hours on 32 A100 GPUs.

\paragraph{Model configurations.}
\Tab{tab:model_config} summarizes the architecture details of all model
variants used in our experiments. The tokenizer consists of a ViT-based encoder
and decoder, offered in Base (B) and Large (L) configurations. The standalone AR
model is a decoder-only Transformer available in three sizes: Base (115M), Large
(305M), and XL (685M). We denote each full system by its tokenizer--AR pairing,
e.g., Prologue B--L refers to a Base tokenizer paired with a Large AR model.

\vspace{10pt}
\begin{table}[h]
\centering
\small
\caption{Model configurations. The Tokenizer consists of Encoder+Decoder; the AR Model is a decoder-only Transformer.}
\label{tab:model_config}
\begin{tabular}{lcccc}
\toprule
& Params & Layers & Dim & Heads \\
\midrule
\multicolumn{5}{l}{\textit{Tokenizer (Encoder/Decoder)}} \\
B & 86M/86M & 12/12 & 768/768 & 12 \\
L & 86M/305M & 12/24 & 768/1024 & 12/16 \\
\midrule
\multicolumn{5}{l}{\textit{AR Model}} \\
B & 115M & 12 & 768 & 12 \\
L & 305M & 24 & 1024 & 16 \\
XL & 685M & 24 & 2048 & 20 \\
\bottomrule
\end{tabular}
\end{table}

\paragraph{Training details}
\textbf{Stage\,1 (Tokenizer + joint AR)}.
The encoder, decoder, and GAN discriminator are optimized with AdamW~\cite{DBLP:journals/corr/abs-1711-05101}
($\beta_1\!=\!0$, $\beta_2\!=\!0.99$), while the compact AR head uses AdamW with
$\beta_1\!=\!0.9$ and $\beta_2\!=\!0.99$. The initial learning rate is
$1\!\times\!10^{-4}$, with linear warmup followed by linear decay to
$3\!\times\!10^{-5}$ for the Base model and $1\!\times\!10^{-5}$ for the Large
model. We use a batch size of 256 and train for 150 epochs (Base) or 200 epochs
(Large). We use vanilla L2-Normed VQ~\cite{DBLP:conf/iclr/YuLKZPQKXBW22} for visual quantizer and Prob-STE for prologue quantizer. The loss weights are set to $\ell_1\!=\!1$, LPIPS$\,{=}\,1$,
$\text{GAN}_G\!=\!0.1$, $\lambda_\text{CE}\!=\!3$, and commitment$\,{=}\,1$.
No weight decay is applied to the encoder and decoder, while the compact AR uses
a weight decay of $3\!\times\!10^{-2}$. Data augmentation consists of random
cropping and horizontal flipping. \textbf{Stage\,2 (Standalone AR training)}.
The AR model paired with the Base tokenizer is trained with AdamW
($\beta_1\!=\!0.9$, $\beta_2\!=\!0.96$) at a learning rate of
$2\!\times\!10^{-4}$ with linear decay to $5\!\times\!10^{-5}$, a batch size of
512, for 400 epochs. When paired with the Large tokenizer, we increase the
learning rate to $4\!\times\!10^{-4}$ with linear decay to
$1\!\times\!10^{-5}$, the batch size to 2048, and train for 800 epochs. Both
configurations use a dropout rate of 0.1, an embedding dropout rate of 0.1,
gradient clipping at 1.0, and a weight decay of $3\!\times\!10^{-2}$. For
Prologue L--XL, we apply a weight decay of $5\!\times\!10^{-2}$ and extend
training by an additional 150 epochs, as we observe that the AR model does not
fully converge even after 800 epochs of training. The class label is dropped with
10\% probability for classifier-free guidance. Data augmentation uses ten-crop.

\paragraph{Prob-STE details.}\label{app:prob-ste}
Given the Encoder output $\mathbf{h}_p$ and codebook vectors $\{\mathbf{c}_i\}$, Prob-STE computes the soft assignment $p_i = \exp(\mathbf{h}_p \cdot \mathbf{c}_i / \tau) / \sum_j \exp(\mathbf{h}_p \cdot \mathbf{c}_j / \tau)$, where $\tau$ is the temperature. The forward pass uses hard assignment $\hat{p}_i=\mathbf{1}[i=\arg\max_j p_j]$; the backward pass propagates gradients through $\tilde{p}_i = p_i + \text{sg}(\hat{p}_i - p_i)$. The quantized representation is $\tilde{\mathbf{q}}_p = \sum_i \tilde{p}_i \cdot \mathbf{c}_i$.

\begin{figure}[h]
    \centering
    \includegraphics[width=0.95\linewidth]{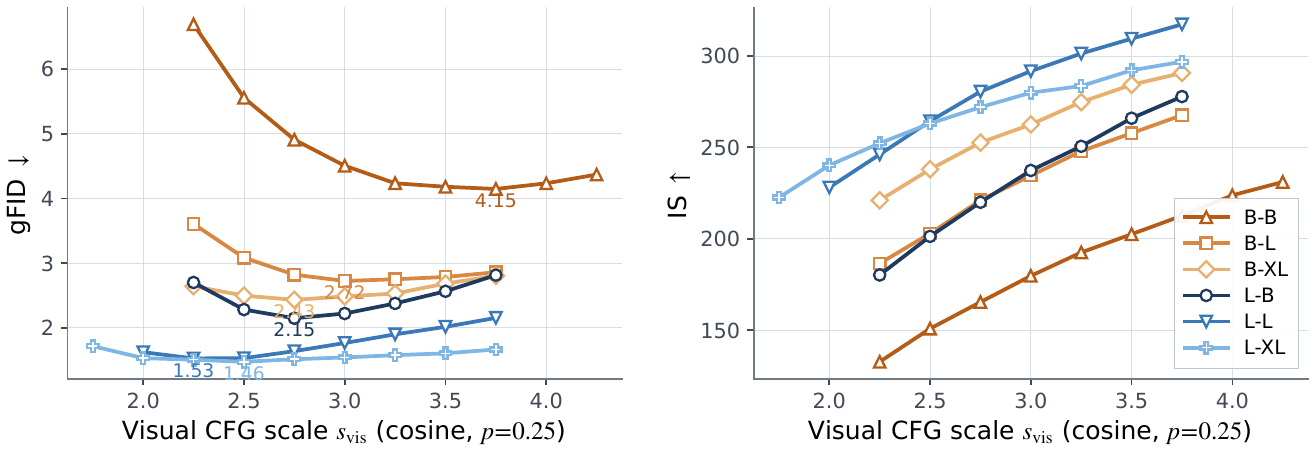}
    \caption{Trend of gFID (left) and IS (right) as the visual CFG scale $s_\mathrm{vis}$ varies (cosine schedule, $p\!=\!0.25$). Each curve fixes $s_\mathrm{pro}$ at the per-variant optimum. gFID forms an interior-minimum U-shape for all six variants while IS increases monotonically, reflecting the fidelity--diversity trade-off.}
    \label{fig:cfg_scale_sweep}
\end{figure}

\paragraph{gFID / IS trend across the CFG scale.}
\Fig{fig:cfg_scale_sweep} shows how gFID and IS vary with the visual CFG scale $s_\mathrm{vis}$ for all six Prologue variants. For each variant we fix the semantic scale $s_\mathrm{pro}$ and the cosine shape parameter $p\!=\!0.25$ at the per-variant optimum and sweep $s_\mathrm{vis}$ over the CFG evaluation set. Every variant exhibits the same trade-off: gFID traces a clean U-shape with an interior minimum, while IS increases monotonically with $s_\mathrm{vis}$---stronger visual guidance yields more class-confident samples at the cost of reduced diversity, and the best gFID is reached before IS saturates. The CFG configurations used for the reported models are listed in \Tab{tab:sampling}.

\vspace{10pt}

\paragraph{Detailed CFG configuration.}
We provide the detailed sampling configurations---including classifier-free
guidance scales and temperature settings---used to obtain all gFID values
reported in this paper in \Tab{tab:sampling}.
\begin{table}[th]
\centering
\caption{Sampling configurations used for the reported gFID.}
\label{tab:sampling}
\resizebox{\textwidth}{!}{%
\begin{tabular}{lllll}
\toprule
Model & Mode & Prologue CFG & Visual CFG & Temperature \\
\midrule
1D Tokenizer & CFG & -- & cosine, $p=1.75$, scale $=19$ &$T=1.0$ \\
1D Tokenizer & w/o CFG & -- & -- & $T=0.85$ \\
2D Tokenizer & CFG & -- & cosine, $p=1.5$, scale $=10$ &$T=1.0$ \\
2D Tokenizer & w/o CFG & -- & -- & $T=0.85$ \\
2D Tokenizer L--B & CFG & -- & cosine, $p=1.75$, scale $=5$ &$T=1.0$ \\
2D Tokenizer L--B & w/o CFG & -- & -- & $T=0.9$ \\
Prologue B--B & CFG & constant scale $=0.7$ & cosine, $p=0.2$, scale $=3.75$ &$T=1.0$ \\
Prologue B--B & w/o CFG & -- & -- & $T_\text{prologue}=0.7$,$T_\text{visual}=0.9$ \\
Prologue-Post & CFG & constant scale $=0.6$ & cosine, $p=0.25$, scale $=3.75$ &$T=1.0$ \\
Prologue-Post & w/o CFG & -- & -- & $T_\text{prologue}=0.8$,$T_\text{visual}=0.8$ \\
Prologue-OneStage & CFG & constant scale $=0.8$ & cosine, $p=0.25$, scale $=3.75$ &$T=1.0$ \\
Prologue-OneStage & w/o CFG & -- & -- & $T_\text{prologue}=0.8$,$T_\text{visual}=0.8$ \\
Prologue B--L & CFG & constant scale $=0.8$ & cosine, $p=0.225$, scale $=3.0$ &$T=1.0$ \\
Prologue B--L & w/o CFG & -- & -- & $T_\text{prologue}=0.7$,$T_\text{visual}=0.9$ \\
Prologue B--XL & CFG & constant scale $=0.8$ & cosine, $p=0.25$, scale $=2.75$ &$T=1.0$ \\
Prologue B--XL & w/o CFG & -- & -- & $T_\text{prologue}=0.8$,$T_\text{visual}=0.9$ \\
Prologue L--B / ConvNeXt-S 4K & CFG & constant scale $=0.65$ & cosine, $p=0.25$, scale $=2.75$ &$T=1.0$ \\
Prologue L--B / ConvNeXt-S 4K & w/o CFG & -- & -- & $T_\text{prologue}=0.8$,$T_\text{visual}=0.95$ \\
Prologue L--L & CFG & constant scale $=0.7$ & cosine, $p=0.25$, scale $=2.5$ &$T=1.0$ \\
Prologue L--L & w/o CFG & -- & -- & $T_\text{prologue}=0.95$,$T_\text{visual}=0.9$ \\
Prologue L--XL & CFG & constant scale $=0.7$ & cosine, $p=0.225$, scale $=2.25$ &$T=1.0$ \\
Prologue L--XL & w/o CFG & -- & -- & $T_\text{prologue}=1.0$,$T_\text{visual}=0.9$ \\
Prologue L--B / VGG16 4K & CFG & constant scale $=0.7$ & cosine, $p=0.25$, scale $=3.75$ &$T=1.0$ \\
Prologue L--B / VGG16 4K & w/o CFG & -- & -- & $T_\text{prologue}=0.8$,$T_\text{visual}=0.85$ \\
Prologue L--B / ConvNeXt-S 16K & CFG & constant scale $=0.65$ & cosine, $p=0.25$, scale $=3.25$ &$T=1.0$ \\
Prologue L--B / ConvNeXt-S 16K & w/o CFG & -- & -- & $T_\text{prologue}=0.8$,$T_\text{visual}=0.95$ \\
Prologue codebook 256 & CFG & constant scale $=0.7$ & cosine, $p=0.25$, scale $=3.25$ &$T=1.0$ \\
Prologue codebook 256 & w/o CFG & -- & -- & $T_\text{prologue}=0.8$,$T_\text{visual}=0.9$ \\
Prologue codebook 4096 & CFG & constant scale $=0.8$ & cosine, $p=0.25$, scale $=3.5$ &$T=1.0$ \\
Prologue codebook 4096 & w/o CFG & -- & -- & $T_\text{prologue}=0.8$,$T_\text{visual}=0.9$ \\
Prologue token count $K=4$ & CFG & constant scale $=0.6$ & cosine, $p=0.25$, scale $=2.5$ &$T=1.0$ \\
Prologue token count $K=4$ & w/o CFG & -- & -- & $T_\text{prologue}=0.8$,$T_\text{visual}=0.9$ \\
Prologue token count $K=64$ & CFG & constant scale $=0.8$ & cosine, $p=0.25$, scale $=3.75$ &$T=1.0$ \\
Prologue token count $K=64$ & w/o CFG & -- & -- & $T_\text{prologue}=0.8$,$T_\text{visual}=1.0$ \\
Prologue Prob-STE $\tau=0.01$ & CFG & constant scale $=0.65$ & cosine, $p=0.25$, scale $=2.5$ &$T=1.0$ \\
Prologue Prob-STE $\tau=0.01$ & w/o CFG & -- & -- & $T_\text{prologue}=0.8$,$T_\text{visual}=0.9$ \\
Prologue Prob-STE $\tau=1.0$ & CFG & constant scale $=0.8$ & cosine, $p=0.25$, scale $=3.75$ &$T=1.0$ \\
Prologue Prob-STE $\tau=1.0$ & w/o CFG & -- & -- & $T_\text{prologue}=0.8$,$T_\text{visual}=0.85$ \\
Prologue Visual dropout $p=0.0$ & CFG & constant scale $=0.65$ & cosine, $p=0.25$, scale $=3.5$ &$T=1.0$ \\
Prologue Visual dropout $p=0.0$ & w/o CFG & -- & -- & $T_\text{prologue}=0.8$,$T_\text{visual}=0.85$ \\
Prologue Visual dropout $p=1.0$ & CFG & constant scale $=0.75$ & cosine, $p=0.25$, scale $=3.25$ &$T=1.0$ \\
Prologue Visual dropout $p=1.0$ & w/o CFG & -- & -- & $T_\text{prologue}=0.8$,$T_\text{visual}=0.9$ \\
Prologue $\lambda=1.0$ & CFG & constant scale $=0.65$ & cosine, $p=0.25$, scale $=3.75$ &$T=1.0$ \\
Prologue $\lambda=1.0$ & w/o CFG & -- & -- & $T_\text{prologue}=0.8$,$T_\text{visual}=0.85$ \\
Prologue $\lambda=3.0$ & CFG & constant scale $=0.7$ & cosine, $p=0.2$, scale $=3.75$ &$T=1.0$ \\
Prologue $\lambda=3.0$ & w/o CFG & -- & -- & $T_\text{prologue}=0.7$,$T_\text{visual}=0.9$ \\
Prologue $\lambda=6.0$ & CFG & constant scale $=0.75$ & cosine, $p=0.25$, scale $=3.25$ &$T=1.0$ \\
Prologue $\lambda=6.0$ & w/o CFG & -- & -- & $T_\text{prologue}=0.8$,$T_\text{visual}=0.9$ \\
\bottomrule
\end{tabular}%
}
\end{table}

\section{Additional Visualization and Results.}
\begin{figure}[th]
    \centering
    \includegraphics[width=\linewidth]{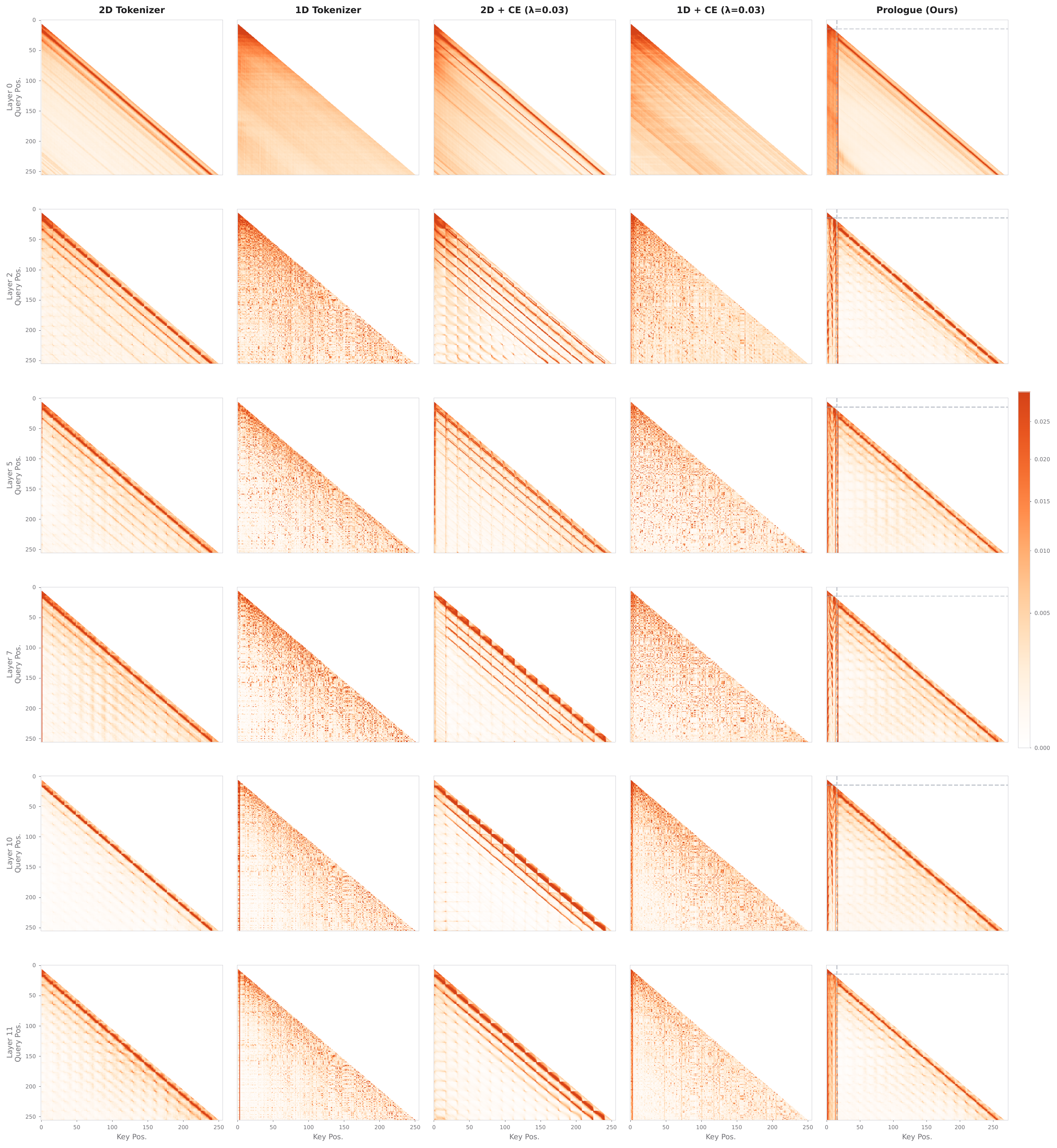}
    \caption{AR attention patterns (Layers 0/2/5/7/10/11; self-attention diagonal and first token masked). In shallow layers (0--2), all models show a uniform and local attention pattern, though the Prologue column is faintly visible. In middle layers (5--7), the Prologue column is most prominent, and pattern differences are most pronounced. In deep layers (10--11), the Prologue column persists; visual tokens attend to the prologue prefix across all depths.}
    \label{fig:attn_all_layers}
\end{figure}

\begin{figure}[h]
    \centering
    \includegraphics[width=\linewidth]{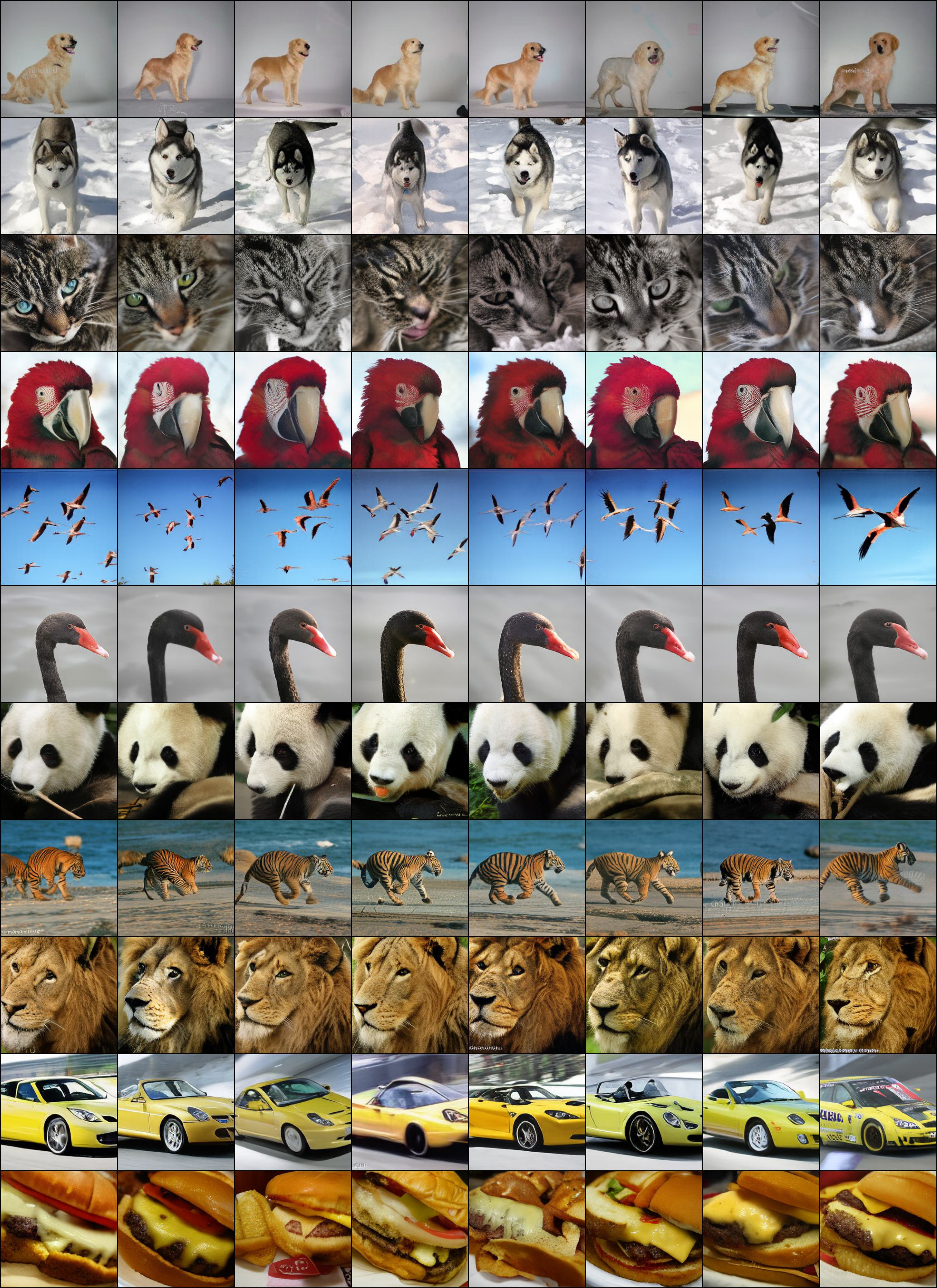}
    \caption{Prologue L--XL generation with \textbf{fixed prologue tokens} (11 classes, 8 visual resamples per class). Within each row, all 16 prologue tokens are held constant and only visual tokens are resampled. Class identity and global layout remain consistent across columns; fine-grained texture and details vary, confirming the semantic-visual disentanglement.}
    \label{fig:appendix_semantic_fix}
\end{figure}

\begin{figure}[h]
    \centering
    \includegraphics[width=\linewidth]{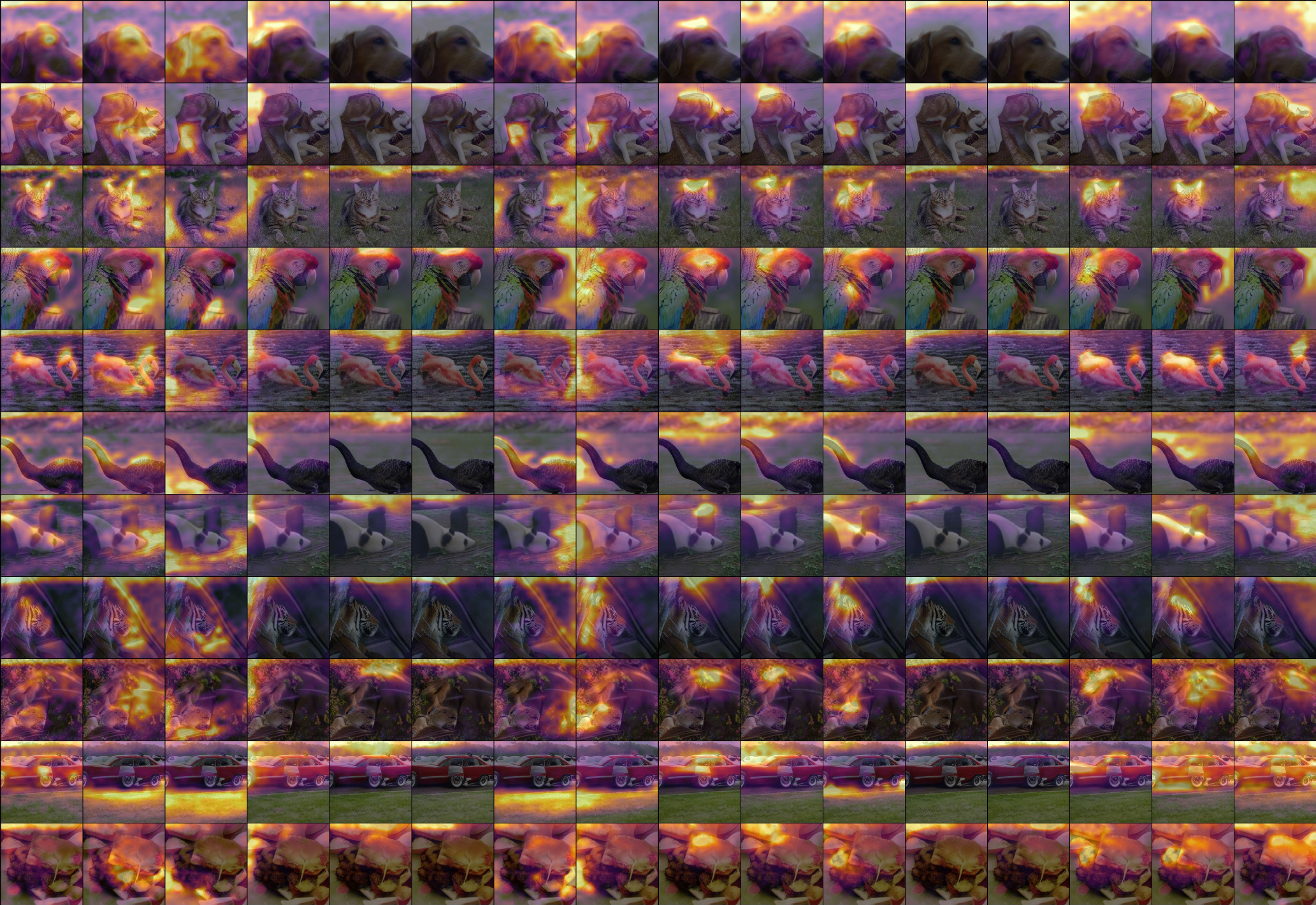}
    \caption{Attention heatmaps from each prologue token to visual tokens (averaged from the last 8 layers and heads). The prologue tokens show rich semantic structure that provides information to visual tokens for objects and background at different regions, indicating a learned spatial division of labor.}
    \label{fig:appendix_ar_attn}
\end{figure}

\begin{figure*}[t]
    \centering
    \includegraphics[width=\linewidth]{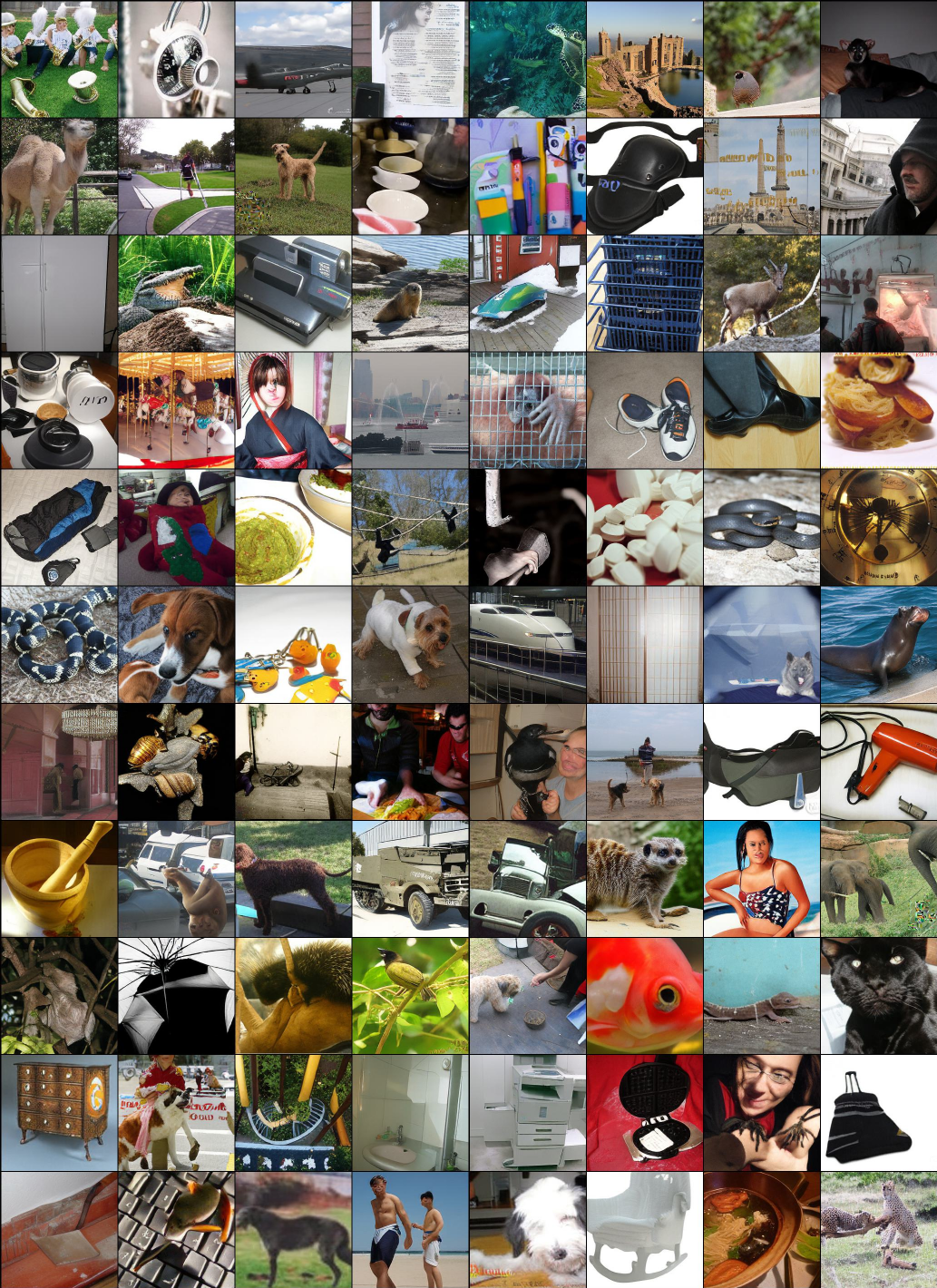}
    \caption{Uncurated samples from Prologue L--XL. Each grid contains 88 randomly selected ImageNet classes with no class repeated across the five grids. Sampling uses prologue CFG $=0.7$ and visual CFG $\mathrm{cosine}(\mathrm{scale}=2.25, p=0.225)$. The model achieves gFID $=1.46$ and IS $=257.66$.}
    \label{fig:uncurated-lxl-00}
\end{figure*}

\begin{figure*}[t]
    \centering
    \includegraphics[width=\linewidth]{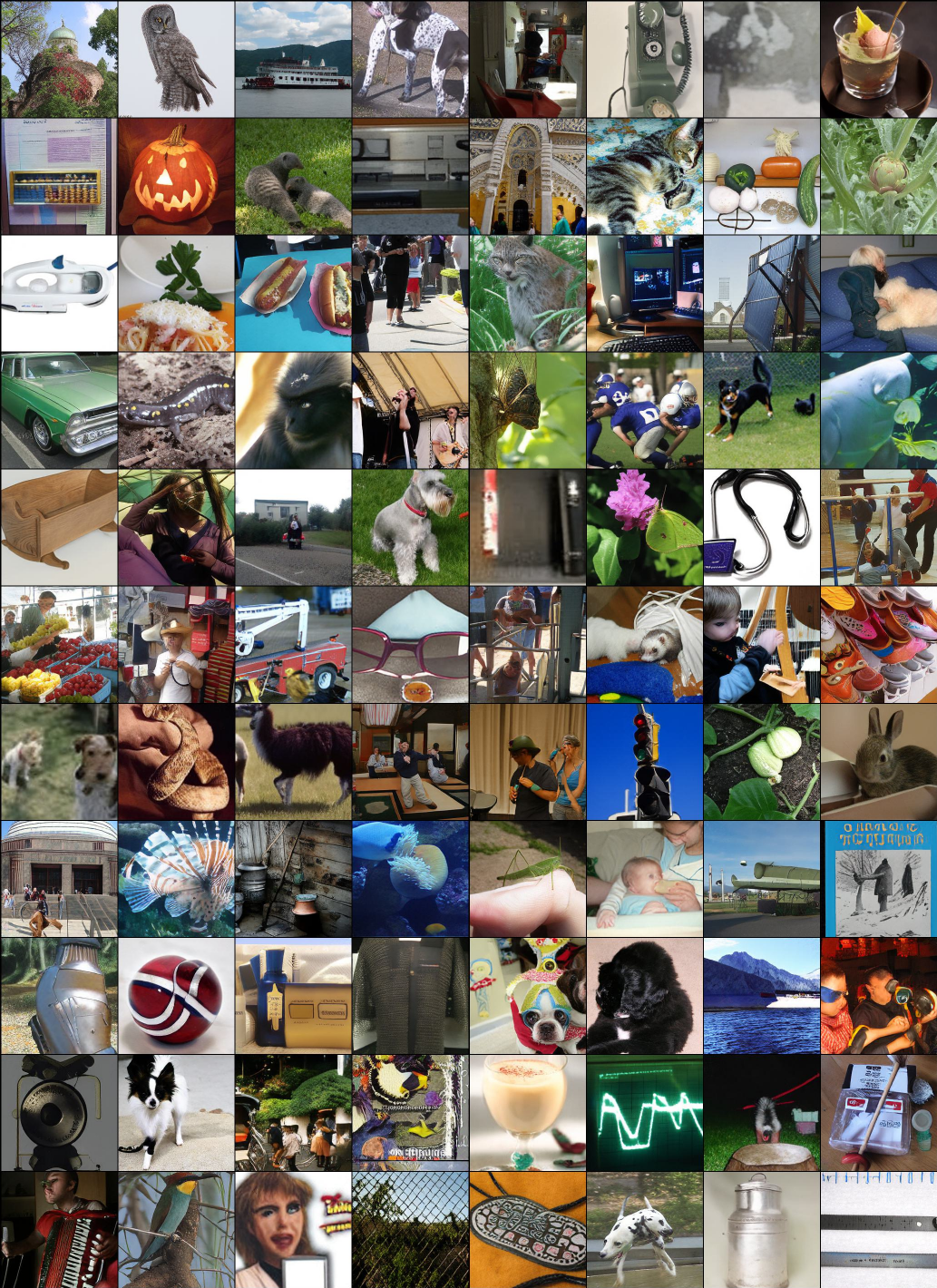}
    \caption{Uncurated samples from Prologue L--XL under the same setting as Fig.~\ref{fig:uncurated-lxl-00}.}
    \label{fig:uncurated-lxl-01}
\end{figure*}

\begin{figure*}[t]
    \centering
    \includegraphics[width=\linewidth]{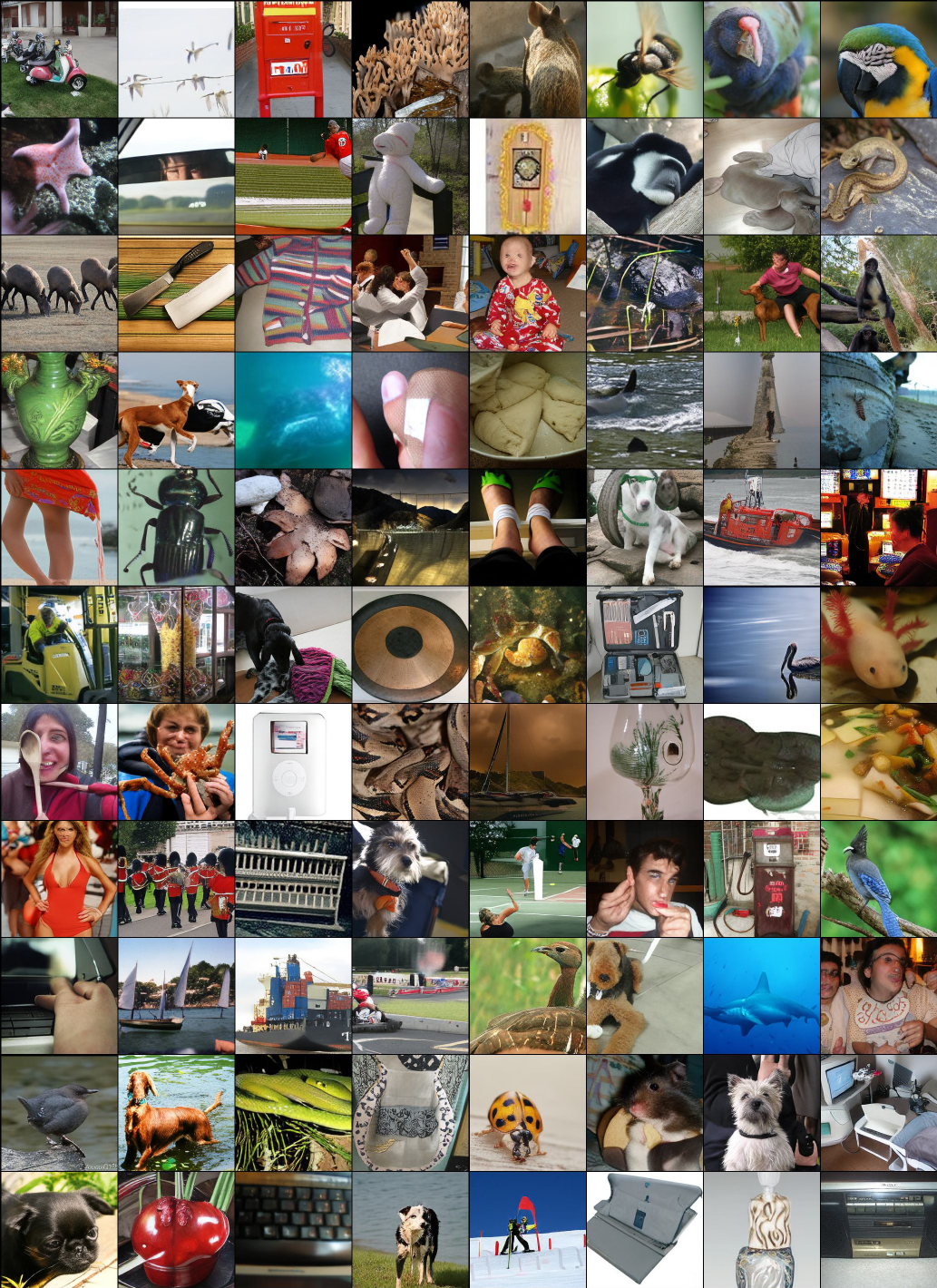}
    \caption{Uncurated samples from Prologue L--XL under the same setting as Fig.~\ref{fig:uncurated-lxl-00}.}
    \label{fig:uncurated-lxl-02}
\end{figure*}

\begin{figure*}[t]
    \centering
    \includegraphics[width=\linewidth]{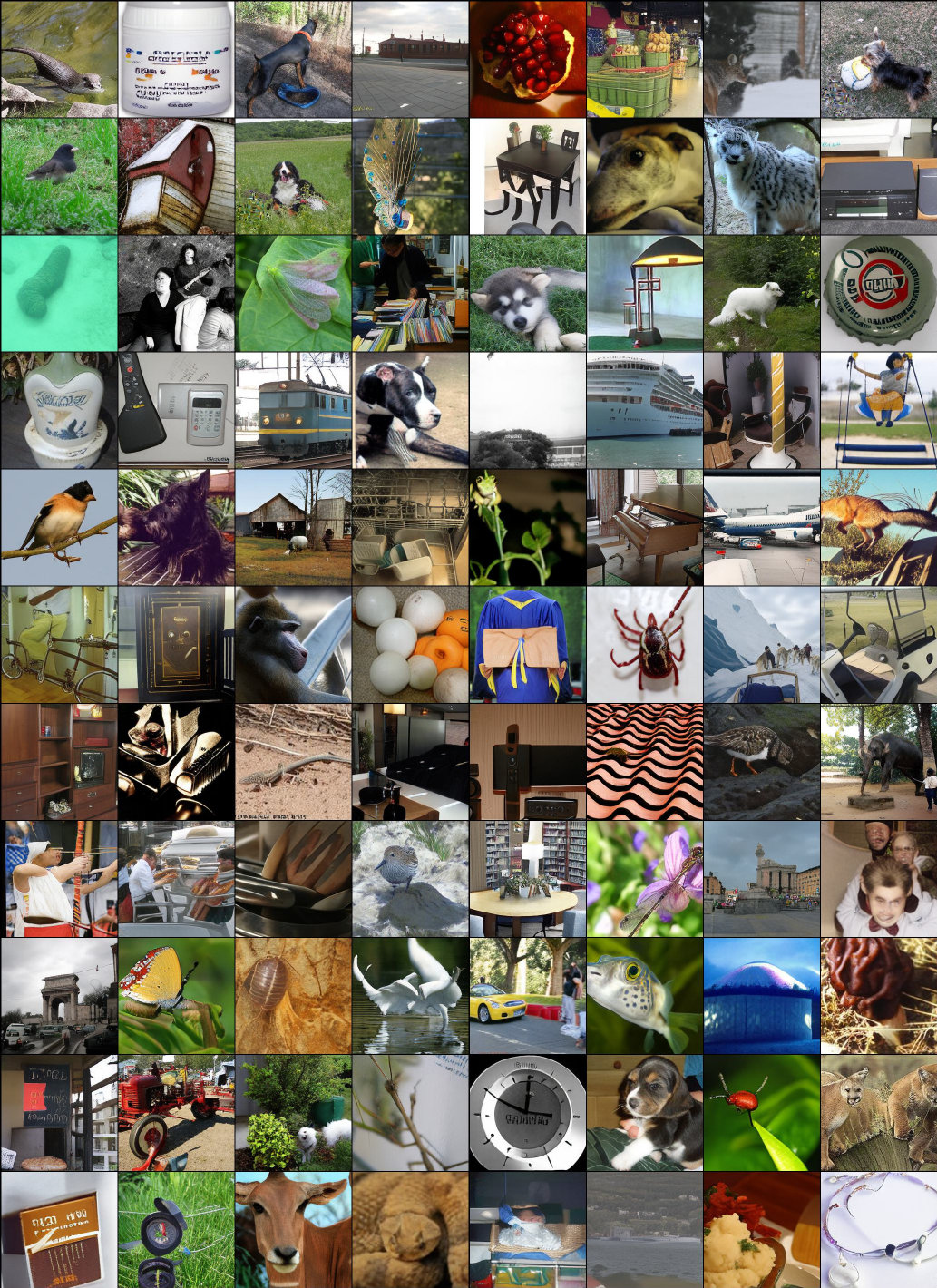}
    \caption{Uncurated samples from Prologue L--XL under the same setting as Fig.~\ref{fig:uncurated-lxl-00}.}
    \label{fig:uncurated-lxl-03}
\end{figure*}

\begin{figure*}[t]
    \centering
    \includegraphics[width=\linewidth]{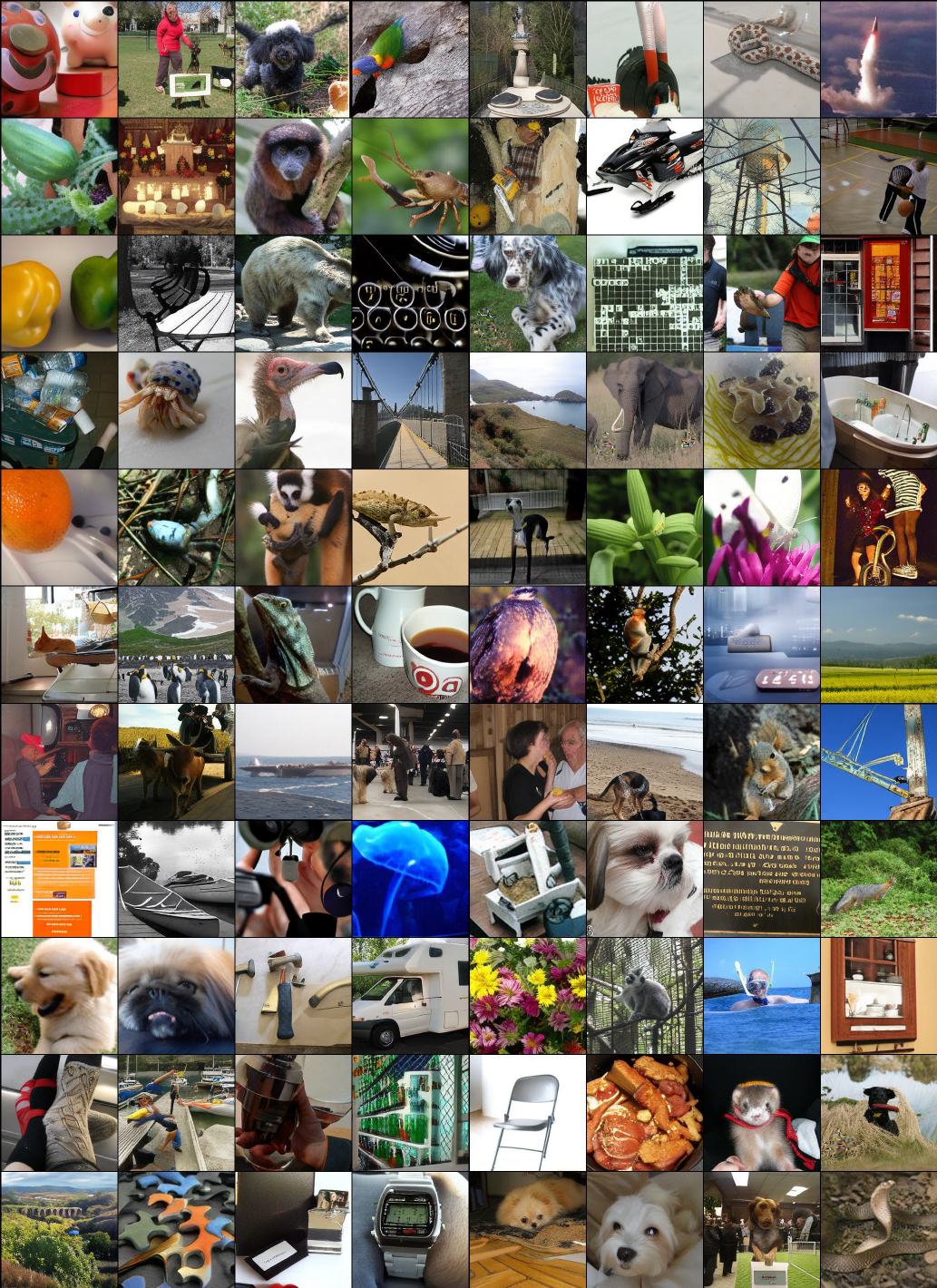}
    \caption{Uncurated samples from Prologue L--XL under the same setting as Fig.~\ref{fig:uncurated-lxl-00}.}
    \label{fig:uncurated-lxl-04}
\end{figure*}

\begin{figure*}[t]
    \centering
    \includegraphics[width=\linewidth]{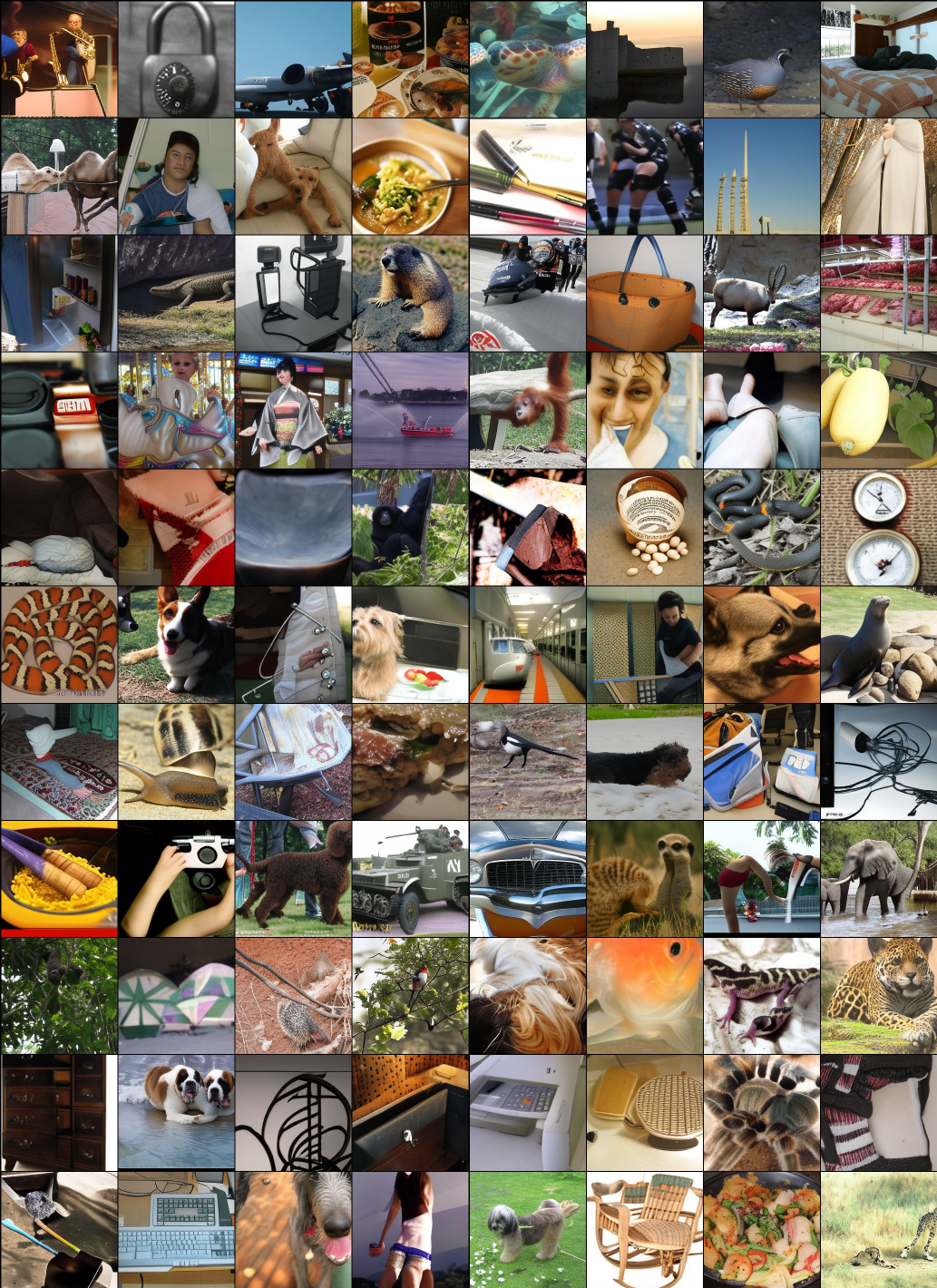}
    \caption{Uncurated samples from the 2D Tokenizer B--B model. Each grid contains 88 randomly selected ImageNet classes with no class repeated across the five grids. Sampling uses CFG $\mathrm{cosine}(\mathrm{scale}=3.75, p=0.2)$. The model achieves gFID $=5.02$.}
    \label{fig:uncurated-bb-00}
\end{figure*}

\begin{figure*}[t]
    \centering
    \includegraphics[width=\linewidth]{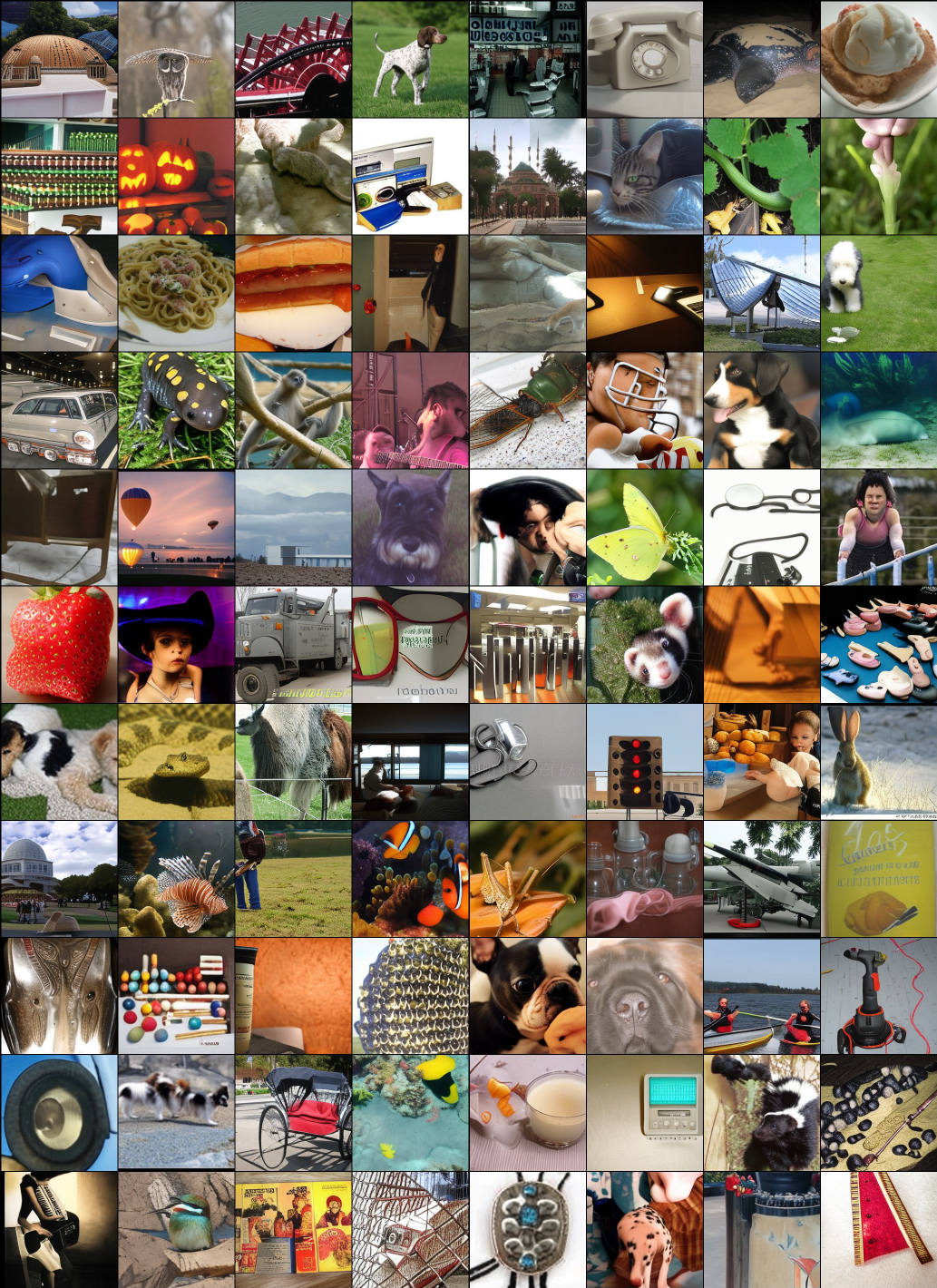}
    \caption{Uncurated samples from the 2D Tokenizer B--B model under the same setting as Fig.~\ref{fig:uncurated-bb-00}.}
    \label{fig:uncurated-bb-01}
\end{figure*}

\begin{figure*}[t]
    \centering
    \includegraphics[width=\linewidth]{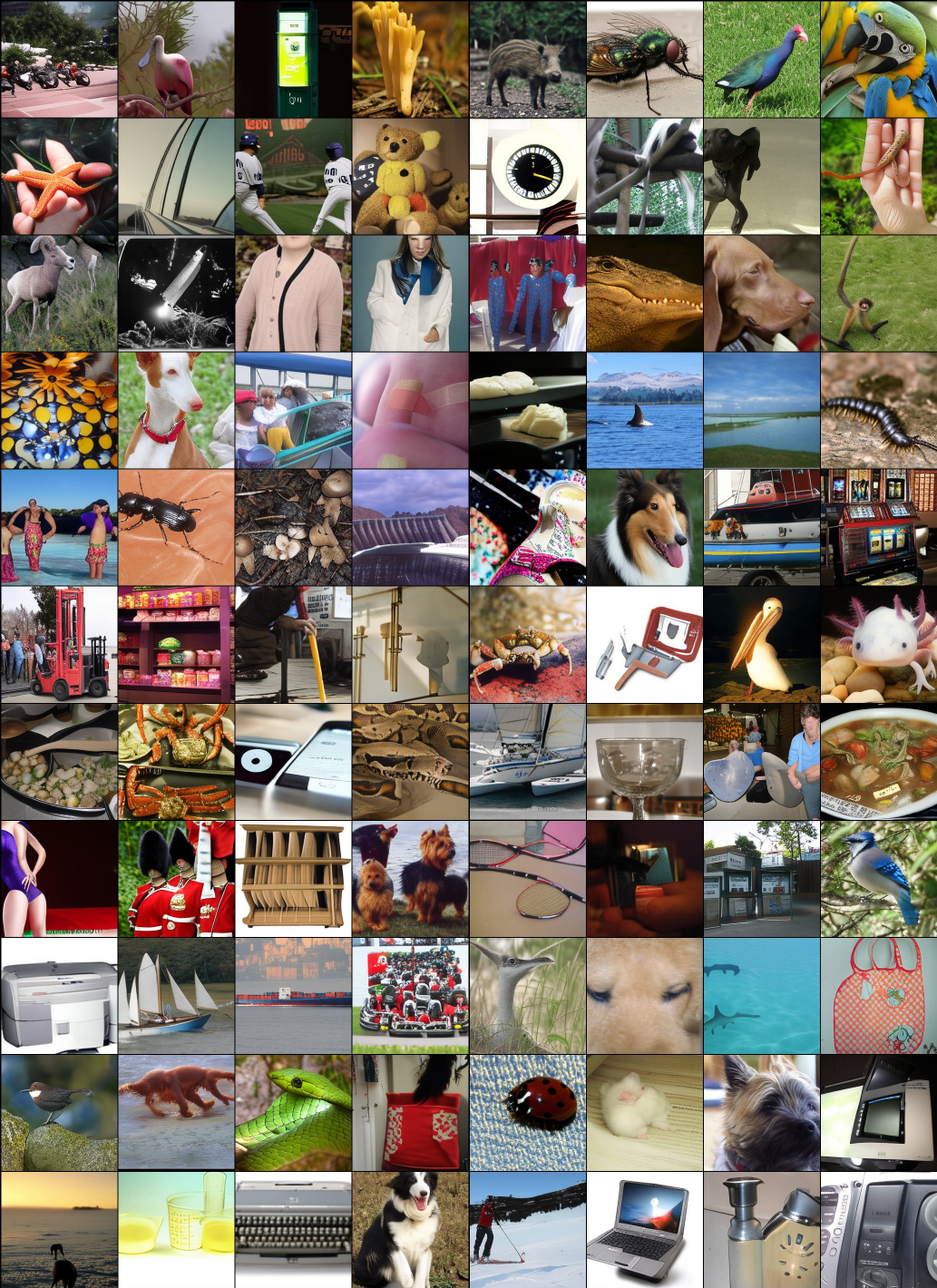}
    \caption{Uncurated samples from the 2D Tokenizer B--B model under the same setting as Fig.~\ref{fig:uncurated-bb-00}.}
    \label{fig:uncurated-bb-02}
\end{figure*}

\begin{figure*}[t]
    \centering
    \includegraphics[width=\linewidth]{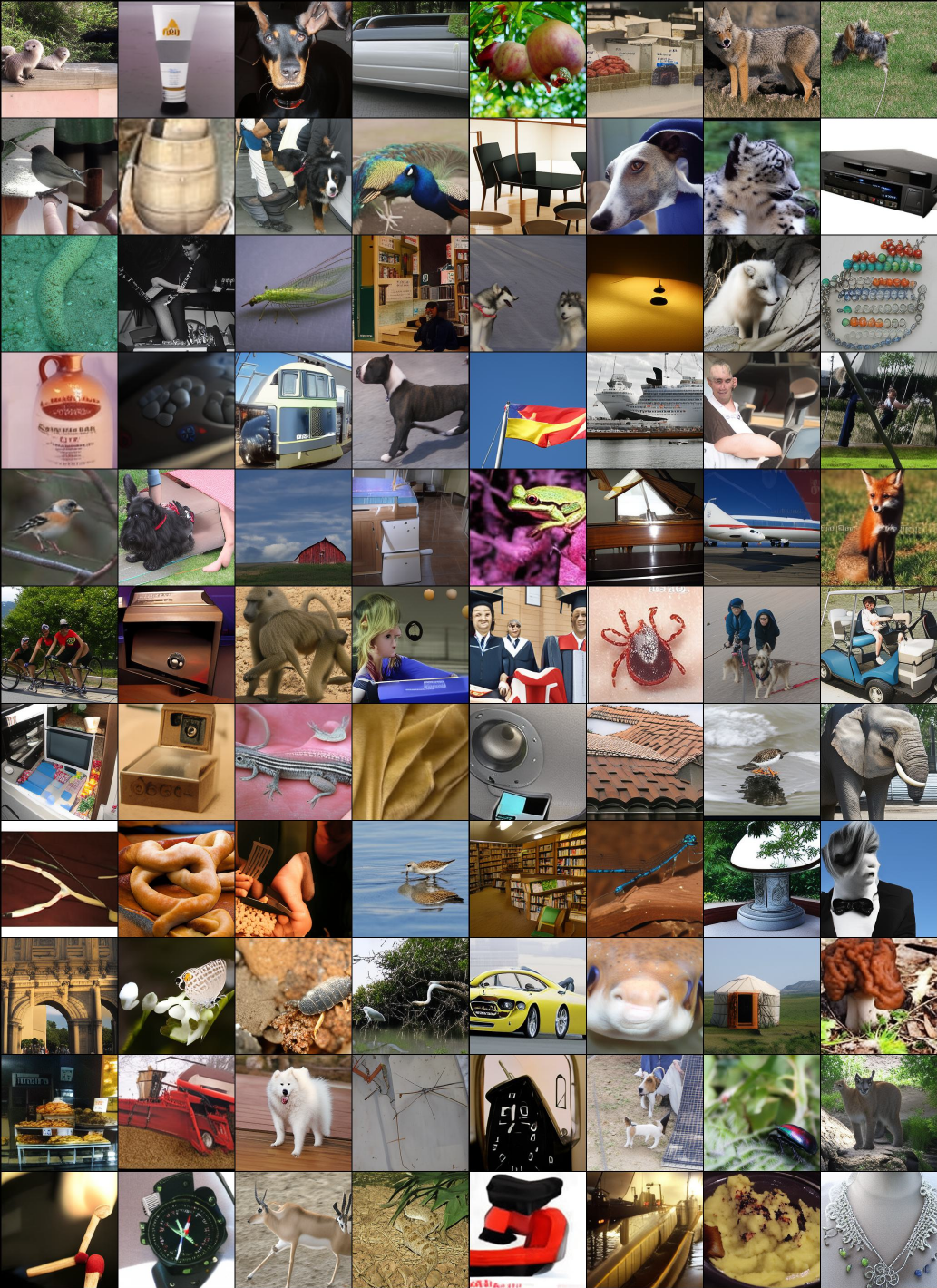}
    \caption{Uncurated samples from the 2D Tokenizer B--B model under the same setting as Fig.~\ref{fig:uncurated-bb-00}.}
    \label{fig:uncurated-bb-03}
\end{figure*}

\begin{figure*}[t]
    \centering
    \includegraphics[width=\linewidth]{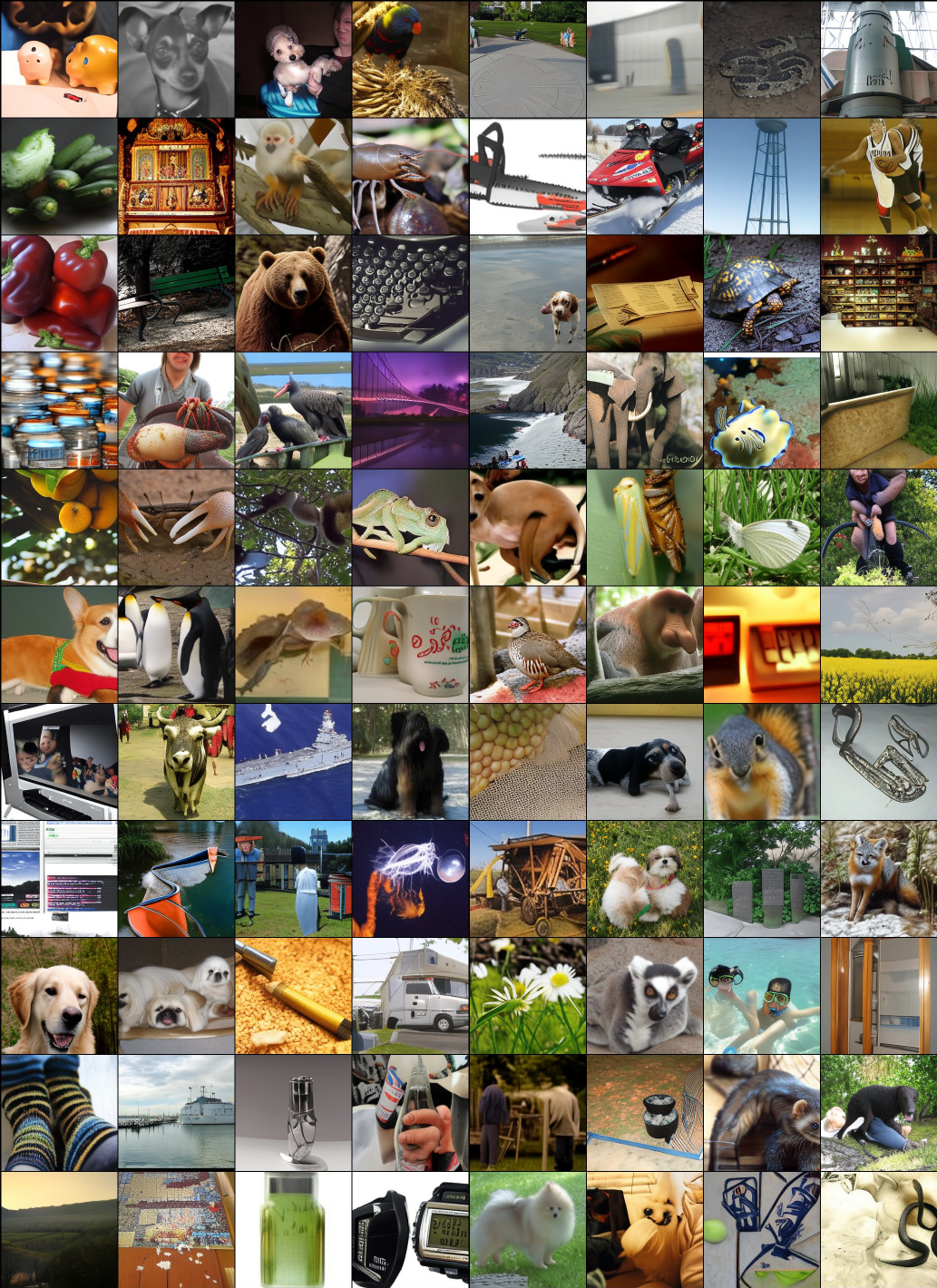}
    \caption{Uncurated samples from the 2D Tokenizer B--B model under the same setting as Fig.~\ref{fig:uncurated-bb-00}.}
    \label{fig:uncurated-bb-04}
\end{figure*}

This appendix provides additional qualitative results.
Fig.~\ref{fig:attn_all_layers} visualizes AR attention patterns across layers.
Fig.~\ref{fig:appendix_semantic_fix} shows semantic-fix generations, where prologue tokens are held fixed while visual tokens are resampled.
Fig.~\ref{fig:appendix_ar_attn} presents per-prologue-token attention heatmaps averaged over the last layers.
Finally, Figs.~\ref{fig:uncurated-lxl-00}--\ref{fig:uncurated-lxl-04} and Figs.~\ref{fig:uncurated-bb-00}--\ref{fig:uncurated-bb-04} show uncurated random samples from Prologue L--XL and the 2D Tokenizer B--B baseline, respectively, generated without manual selection using non-overlapping random ImageNet classes.

\end{document}